\documentclass[final]{cvpr}

\usepackage{times}
\usepackage{epsfig}
\usepackage{graphicx}
\usepackage{amsmath}
\usepackage{amssymb}
\usepackage{multirow}
\usepackage{subcaption}
\usepackage{multicol}
\usepackage{stfloats}
\usepackage{makecell}
\usepackage{graphbox}

\usepackage[pagebackref=true,breaklinks=true,colorlinks,bookmarks=false]{hyperref}

\def\cvprPaperID{10132} 
\def\confYear{CVPR 2021}

\begin{document}

\title{Decoder Modulation for Indoor Depth Completion}

\author{Senushkin Dmitry\\
Samsung AI Center, Moscow\\
{\tt\small d.senushkin@partner.samsung.com}
\and
Romanov Mikhail\\
Samsung AI Center, Moscow\\
{\tt\small m.romanov@samsung.com}

\and
Belikov Ilia\\
Samsung AI Center, Moscow\\
{\tt\small ilia.belikov@samsung.com}

\and
Konushin Anton\\
Samsung AI Center, Moscow\\
{\tt\small a.konushin@samsung.com}

\and
Patakin Nikolay\\
Samsung AI Center, Moscow\\
{\tt\small n.patakin@samsung.com}

}

\maketitle

\begin{abstract}
    Depth completion recovers a dense depth map from sensor measurements. Current methods are mostly tailored for very sparse depth measurements from LiDARs in outdoor settings, while for indoor scenes Time-of-Flight (ToF) or structured light sensors are mostly used. These sensors provide semi-dense maps, with dense measurements in some regions and almost empty in others. We propose a new model that takes into account the statistical difference between such regions. Our main contribution is a new decoder modulation branch added to the encoder-decoder architecture. The encoder extracts features from the concatenated RGB image and raw depth. Given the mask of missing values as input, the proposed modulation branch controls the decoding of a dense depth map from these features differently for different regions. This is implemented by modifying the spatial distribution of output signals inside the decoder via Spatially-Adaptive Denormalization (SPADE) blocks. Our second contribution is a novel training strategy that allows us to train on a semi-dense sensor data when the ground truth depth map is not available. Our model achieves the state of the art results on indoor \emph{Matterport3D} dataset~\cite{Matterport3D}. Being designed for semi-dense input depth, our model is still competitive with LiDAR-oriented approaches on the \emph{KITTI} dataset~\cite{uhrig}. Our training strategy significantly improves prediction quality with no dense ground truth available, as validated on the \emph{NYUv2} dataset~\cite{nyuv2}. 

\end{abstract}

\section{Introduction}

    In recent years, depth sensors have become an essential component of many devices, from self-driving cars to smartphones. However, the quality of modern depth sensors is still far from perfect. LiDAR systems provide accurate but spatially sparse measurements while being quite expensive and large. Commodity-grade depth sensors based on the active stereo with structured light (\eg, Microsoft Kinect) or Time-of-Flight (\eg, Microsoft Kinect Azure or depth sensors in many smartphones) provide estimations that are relatively dense but less accurate and within a limited distance range. LiDAR-based sensors are widely used in outdoor environments, especially for self-driving cars, while the other sensors are mainly applicable in an indoor setting. Due to the rapid growth of the self-driving car industry, the majority of recent depth completion methods are mostly focused on outdoor depth completion for LiDAR data~\cite{uhrig, guidenet, cspn}, often overlooking other types of sensors and scenarios. Nevertheless, these sensors are an essential part of many modern devices (such as mobile phones, AR glasses, and others). 
    
    \begin{figure}[t]
      \centering
      \begin{subfigure}[b]{0.49\linewidth}
        \centering
        \includegraphics[width=\linewidth]{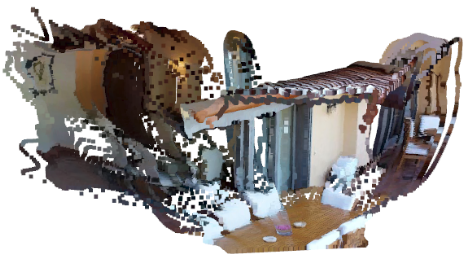}
        \centering
        \includegraphics[width=\linewidth]{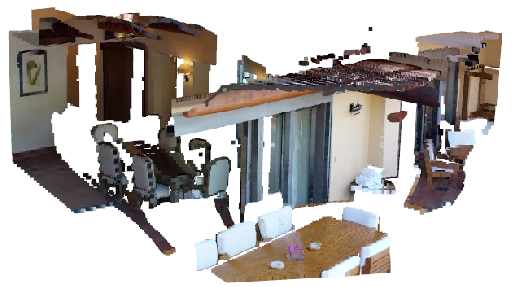}
      \end{subfigure}
      \begin{subfigure}[b]{0.49\linewidth}
        \centering
        \includegraphics[width=\linewidth]{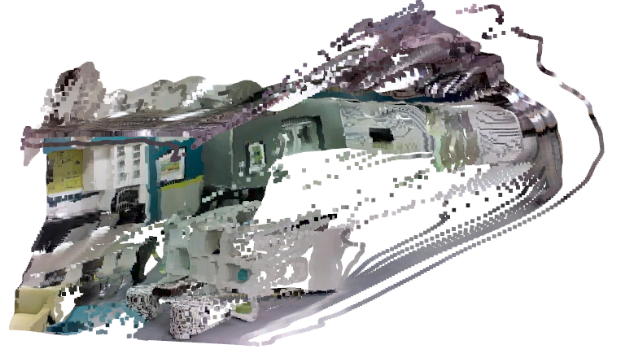}
        \centering
        \includegraphics[width=\linewidth]{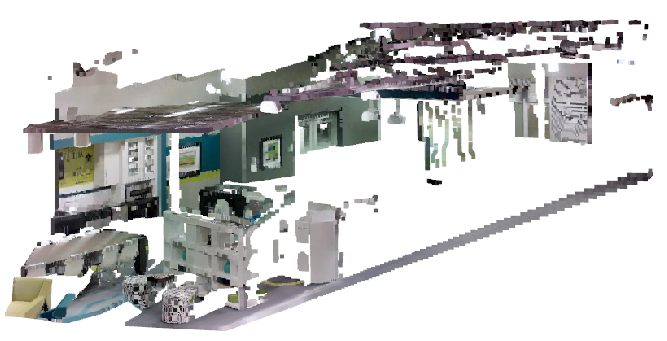}
      \end{subfigure}
      \caption{Point clouds are reconstructed from depth maps predicted by our model (top row) and ground truth (bottom row) taken from the \emph{Matterport3D}~\cite{Matterport3D} test subset. }
      \label{fig:mp3d_point_clouds}
    \end{figure}
    
    LiDAR-oriented methods mainly deal with sparse measurements.
    Applying these methods to depth data captured with semi-dense sensors as-is may be a suboptimal strategy. This kind of transfer requires additional heuristics such as sparse random sampling. The most popular approach~\cite{guidenet, deeplidar, Cheng_2018_ECCV, Ma2017SparseToDense} 
    of training LiDAR-oriented methods on a semi-dense depth map proceeds as follows. First, the gaps in semi-dense depth maps are filled using simple interpolation methods such as bilateral filtering or the approach of~\cite{levin}. Then, some depth points are uniformly sampled from the resulting map. This heuristic approach is used due to the lack of LiDAR data for indoor environments, but such kind of preprocessing suggests that it may be better to use a model originally designed to operate with semi-dense data. Such an approach would take into account the features of semi-dense sensor data and would not require separate heuristics for transfer.

    Inspired by these observations, we present a novel solution for the indoor depth completion from semi-dense depth maps guided by color images. Since sensor data may be present for 60\% of pixels and more, we propose to use a single encoder for the joint RGBD signal. Taking into account the statistical differences between regions with and without depth measurements, we design a decoder modulation branch that takes a mask as input and modifies the distributions of activation maps in the decoder. This modulation mechanism is based on Spatially-Adaptive Denormalization (SPADE) blocks~\cite{spade}.
    Since there are few publicly available datasets with both sensor and dense ground truth depth, we additionally propose a special training strategy for depth completion models that emulates semi-dense sensors and does not require dense depth reconstruction. 
    
    As a result, we offer the following \textbf{contributions}:
    \begin{itemize}
        \item a novel network architecture for indoor depth completion with a decoder modulation branch;
        \item a novel training strategy that emulates semi-dense sensors and does not require dense depth reconstruction;
        \item large-scale experimental validation on real datasets including Matterport3D, ScanNet, NYUv2, and KITTI.
    \end{itemize}
    
    The paper is organized as follows. In Section~\ref{sec:related}, we review related work on depth estimation and dense image labeling. Section~\ref{sec:methods} presents our approach, including the new architecture and training strategies. Section~\ref{sec:experiments} describes the experimental setup, Section~\ref{sec:Matterport3D} presents the results of our experiments, and Section~\ref{sec:concl} concludes the paper.

\section{Related work}\label{sec:related}

    In this section, we review works on several topics related to depth processing for images or works that have served as the original inspiration for our work. Namely, we cover depth estimation, depth completion, and semantic segmentation, a well-studied case of dense image labeling.
    
    \paragraph{Depth Estimation.}
    
    Methods for single view depth estimation based on deep neural networks have significantly evolved in recent years, by now rapidly approaching the accuracy of depth sensors~\cite{casser2018depth, fu2018deep, liang2018learning, luo2016efficient}; some of these methods are able to run in real-time~\cite{icra_2019_fastdepth} or even on embedded platforms~\cite{Ambarella}. However, the acquisition of accurate ground truth depth maps is complicated due to certain limitations of existing depth sensors. To overcome these difficulties, various approaches focusing on data acquisition, data refinement, and the use of additional alternative data sources have been proposed~\cite{DBLP:journals/corr/abs-1804-00607, DBLP:journals/corr/abs-1907-01341}. We also note several recently developed weakly supervised and unsupervised approaches~\cite{Ren_2020_CVPR_Workshops, 8100182}. 
    
    \paragraph{Depth Completion.}
    Pioneering works on depth completion adopted complicated heuristic algorithms for processing raw sensor data. These algorithms were based on compressed sensing~\cite{6126488} or used a combined wavelet-contourlet dictionary~\cite{7055919}.
    Uhrig \etal~\cite{uhrig} were the first to present a successful learnable depth completion method based on convolutional neural networks, developing special sparsity-invariant convolutions to handle sparse inputs. Learnable methods were further improved by image guidance~\cite{cspn, 8757939, Yang_2019_CVPR, 8917294}. Tang \etal~\cite{guidenet} proposed an approach to train content-dependent and spatially-variant kernels for sparse depth features processing. Li \etal~\cite{hourglass} suggested a multi-scale guided cascade hourglass architecture for depth completion. Chen \etal~\cite{uberfusenet} presented a 2D-3D fusion pipeline based on continuous convolutions. Apart from utilizing images, some recently proposed methods make use of surface normals~\cite{deeplidar, Huang_2019, Xu_2019_ICCV, DBLP:journals/corr/abs-1803-09326} and object boundaries~\cite{Huang_2019, DBLP:journals/corr/abs-1803-09326}.
    
    Most of the above-mentioned works focus on LiDAR-based sparse depth completion in outdoor scenarios and report results on the well-known KITTI benchmark~\cite{uhrig}. There are only a few works that consider processing non-LiDAR semi-dense depth data obtained with Kinect sensors. Recently, Zhang \etal~\cite{DBLP:journals/corr/abs-1803-09326} introduced \emph{Matterport3D}, a large-scale RGBD dataset for indoor depth completion, and used it to showcase a custom depth completion method. This method implicitly exploits extra data by using pretrained networks for normal estimation and boundary detection, and the resulting normals and boundaries are used in global optimization. Overall, the complexity of this method strictly limits its practical usage. Huang \etal~\cite{Huang_2019} was the first to outperform the original results on this dataset. Similar to Zhang \etal~\cite{DBLP:journals/corr/abs-1803-09326}, their results were achieved via a complicated multi-stage method that involved resource-intensive preprocessing. Although it does not rely on pretrained backbones, it uses a normal estimation network explicitly trained on external data. In this work, we propose a novel depth completion method that presents strong baseline results while being scalable and straightforward.
    
    \paragraph{Semantic segmentation and dense labeling.}
    
    Since depth completion or depth estimation can be formulated as a dense labeling problem, techniques and architectures that have proven to be effective for other dense labeling tasks might be useful for depth completion as well. Encoder-decoder architectures with skip connections originally developed for semantic segmentation~\cite{unet} have shown themselves to be capable of solving a wide range of tasks. Chen \etal \cite{deeplab} proposed a powerful architecture based on atrous spatial pyramid pooling for semantic segmentation and improved it in further work~\cite{DBLP:journals/corr/ChenPSA17}. Other important approaches include the refinement network~\cite{refinenet} and the pyramid scene parsing network~\cite{pspnet}. At the same time, lightweight networks such as~\cite{lrn-de} capable of running in a resource-constrained device in real-time can be of use in other deep learning-driven applications. Our depth completion network is based on the blocks proposed in~\cite{refinenet, LRN}.

\section{Approach and methods}\label{sec:methods}

    \begin{figure}[t]
    \begin{center}
       \includegraphics[width=1\linewidth]{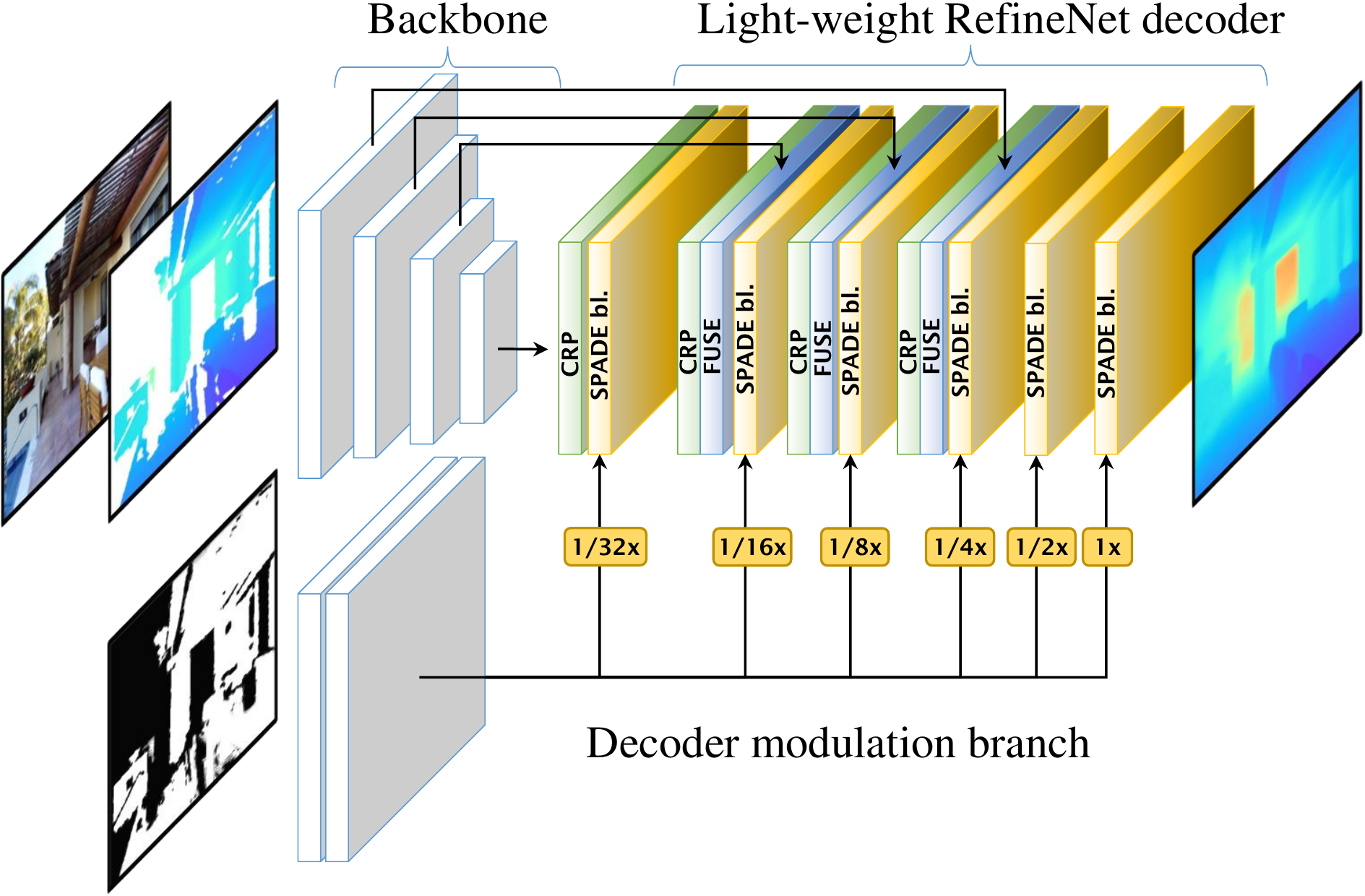}
    \end{center}
    \caption{High-level architecture of the proposed DM-LRN network. Pretrained EfficientNet \cite{EffitientNet} backbone encodes the input RGBD signal. Extracted features are fed into the lightweight RefineNet decoder~\cite{LRN} consisting of chained residual pooling (CRP) blocks and fusion (FUSE) blocks~\cite{LRN}. The decoder modulation branch modifies the spatial distribution of output signals inside the decoder via SPADE blocks~\cite{spade}.}
    \label{fig:arch}
    \end{figure}
    
    
    \paragraph{Architecture overview.}
    The general structure of the proposed architecture is shown in Fig.~\ref{fig:arch}. Our architecture design follows the standard encoder-decoder paradigm with a pretrained backbone network modified for 4D input. In our experiments, we use the EfficientNet family~\cite{EffitientNet} as a backbone. The decoder part is based on a lightweight RefineNet decoder~\cite{LRN} combined with a custom modulation branch described below. The network takes an image, sensor depth, and a mask as input and outputs a completed depth map. No additional data is required.

    \paragraph{Decoder Modulation Branch.}\label{par:decoder_modulation_branch}
    To introduce the decoder modulation branch, let us take a closer look at the forward propagation path of the network. The backbone network generates feature maps from the input RGBD signal. The input signal initially has an inhomogeneous spatial distribution, since a part of the depth data is missing. The signal compression inside a backbone smoothes this inhomogeneity, which works well for small depth gaps. If the depth gaps are too large, the convolutions generate incorrect activations due to the domain shift between RGB and RGBD signals. Aiming to reduce this domain gap, we propose to learn spatially-dependent scale and bias for normalized feature maps inside the decoder part of the architecture. This procedure is called spatially-adaptive denormalization (SPADE) and was first introduced by Park \etal~\cite{spade}.
    
    Let $f^i_{n,c,y,x}$ denote the activation maps of the $i$th layer of the decoder for a batch of $N$ samples with shape $C_i \times H_i \times W_i$, and let $\textbf{m}$ denote a modulation signal. The output value from SPADE $g^i_{n,c,y,x}$ at location $(n \in N, c \in C_i, y \in H_i, x \in W_i)$ is
    $$
        g^i_{n,c,y,x} =  \gamma^i_{n,c,y,x}(\textbf{m}) \frac{f^i_{n,c,y,x} - \mu^i_c}{\sigma^i_c} + \beta^i_{n,c,y,x}(\textbf{m}),
    $$
    where $\mu^i_c = \frac{1}{N_iW_iH_i} \sum_{n,x,y} f^i_{n,c,y,x}$ is the sample mean and $\sigma^i_c= \sqrt{\frac{1}{N_iW_iH_i} \sum_{n,x,y}(f^i_{n,c,y,x} - \mu^i_c)^2 }$ is the sample (biased) standard deviation, and $\gamma^i_{n,c,y,x}$ and $\beta^i_{n,c,y,x}$ are the spatially dependent scale and bias for batch normalization respectively. In our case, the modulation signal $\textbf{m}$ is the input mask of missing depth values. 
    
\begin{figure}[t]
      \centering
      \begin{subfigure}[b]{0.5\linewidth}
        \includegraphics[width=\linewidth]{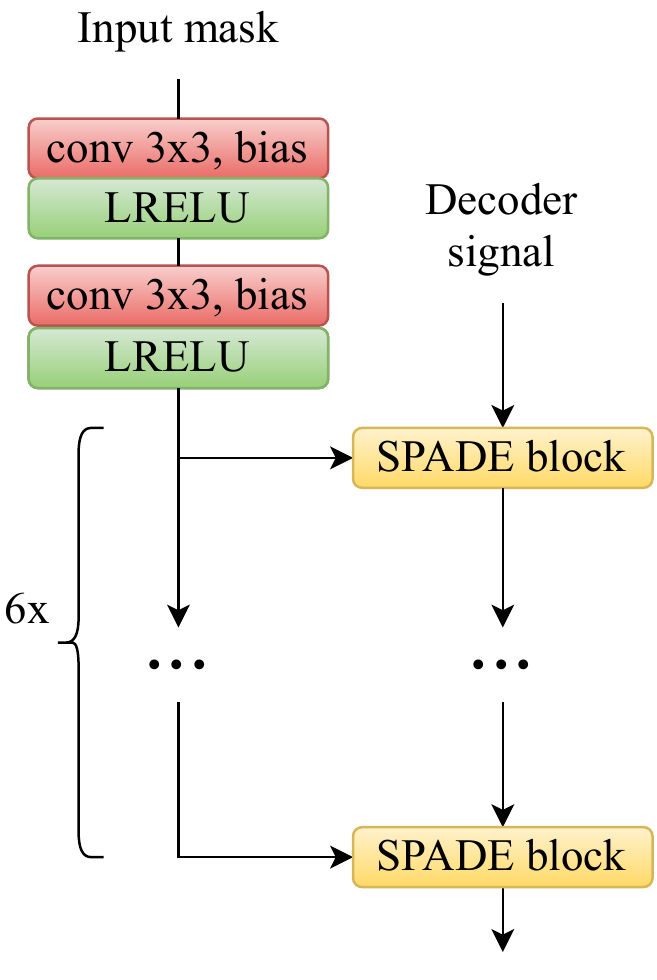}
        \caption{Decoder modulation branch.}
        \label{fig:dm-impl}
      \end{subfigure}
      \begin{subfigure}[b]{0.4\linewidth}
        \centering
        \includegraphics[width=0.7\linewidth]{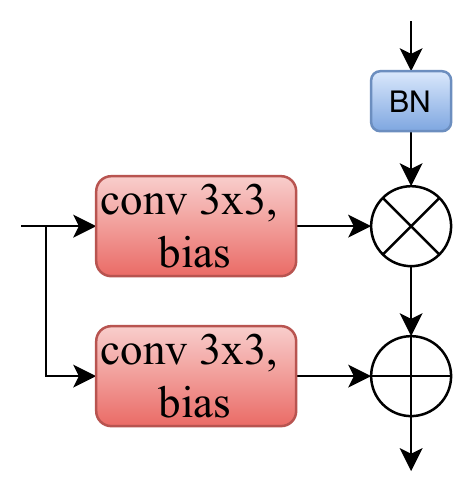}
        \caption{SPADE.}
        \label{fig:spade}
        \centering
        \includegraphics[width=\linewidth]{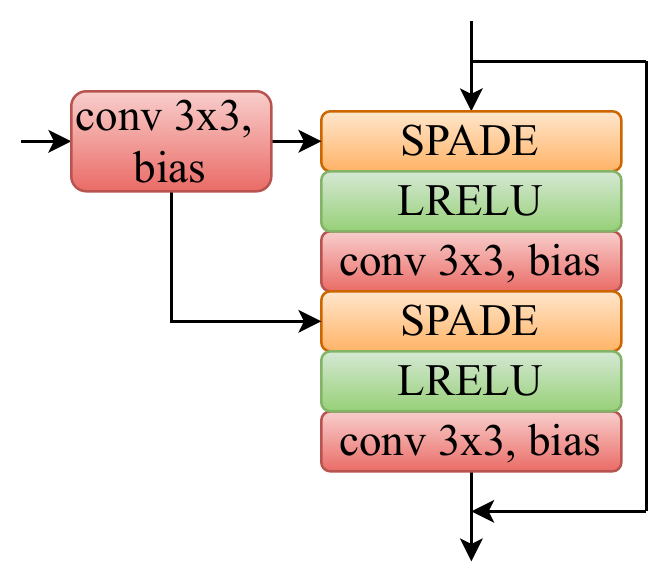}
        \caption{SPADE block.}
        \label{fig:spade-bl}
      \end{subfigure}
      \caption{Architecture of the decoder modulation branch (\ref{fig:dm-impl}). It consists of a simple encoder composed of two biased convolutions with activations and a series of SPADE blocks~(\ref{fig:spade-bl}). These blocks include the SPADE layer (\ref{fig:spade}) that performs modulation. We use LeakyReLU activations, as the modulation should be able to decrease the scale of a signal and move it in the negative direction as well.}
      \label{fig:dm}
    \end{figure}

    Fig.~\ref{fig:dm} illustrates the decoder modulation branch in detail. This subnetwork consists of a simple mask encoder composed of convolutions with bias terms and activations and SPADE blocks that perform modulation. A bias term in the convolutions is necessary to avoid zero signals that can cover a significant part of the input mask.
    
    \paragraph{Training strategy.}
    Existing highly annotated large-scale indoor datasets do not always include both sensor depth data and ground truth depth data \cite{taskonomy2018, nyuv2}, which might be an issue for the development of depth completion models. If the sensor or reconstructed depth is not available, we propose to use specially developed corruption techniques in order to obtain synthetic semi-dense sensor data.
    
\begin{figure}[t]
      \centering
      \begin{subfigure}[b]{0.32\linewidth}
        \centering
        \includegraphics[width=\linewidth]{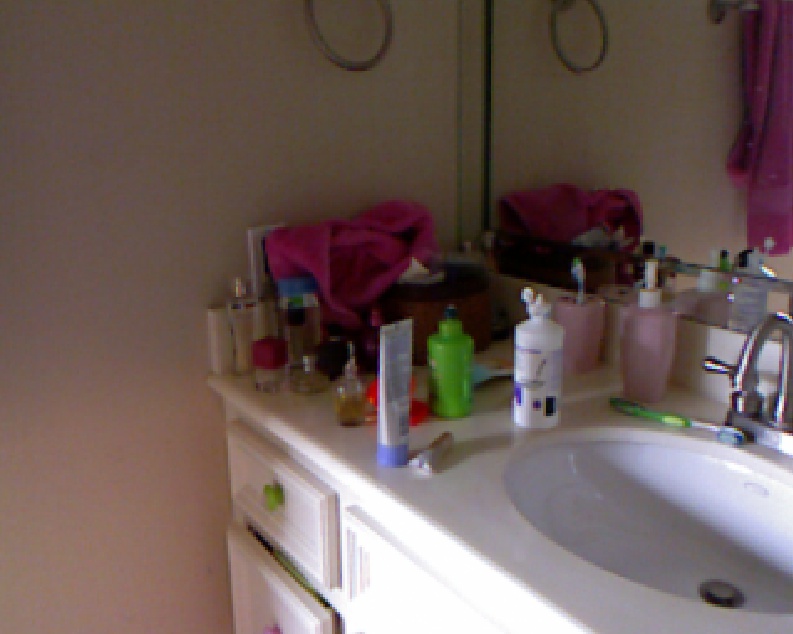}
        \caption{RGB}
        \label{fig:rgb}
        \centering
        \includegraphics[width=\linewidth]{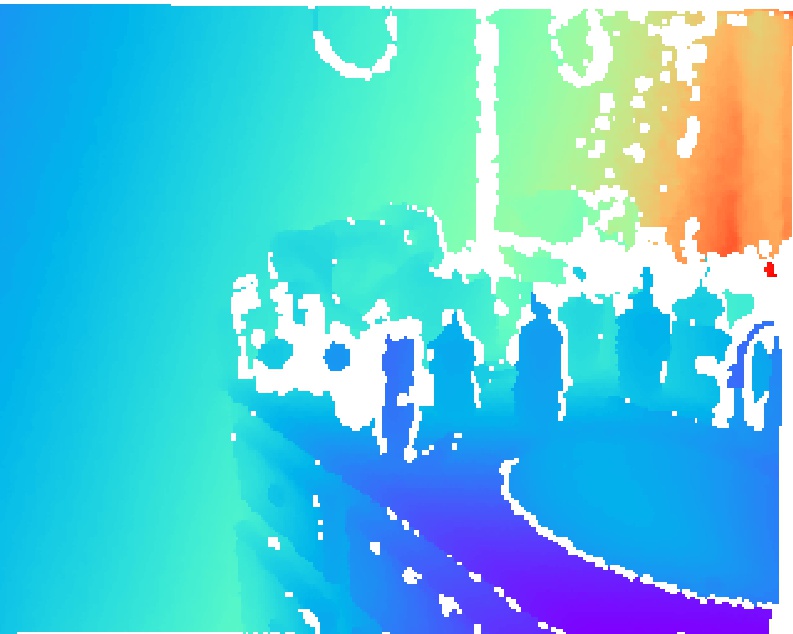}
        \caption{Initial real sensor}
        \label{fig:gt}
      \end{subfigure}
      \begin{subfigure}[b]{0.32\linewidth}
        \centering
        \includegraphics[width=\linewidth]{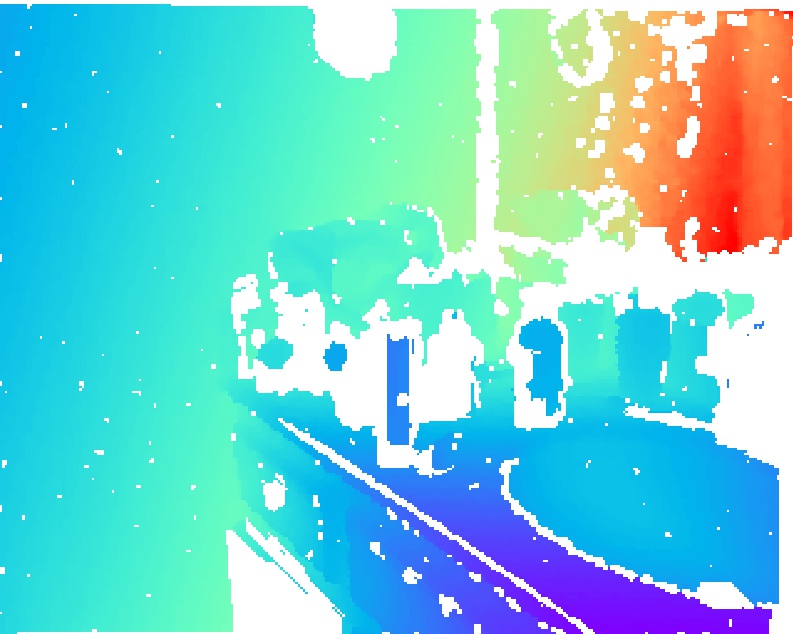}
        \caption{Graph-based \cite{felzenszwalb}}
        \label{fig:felz}
        \centering
        \includegraphics[width=\linewidth]{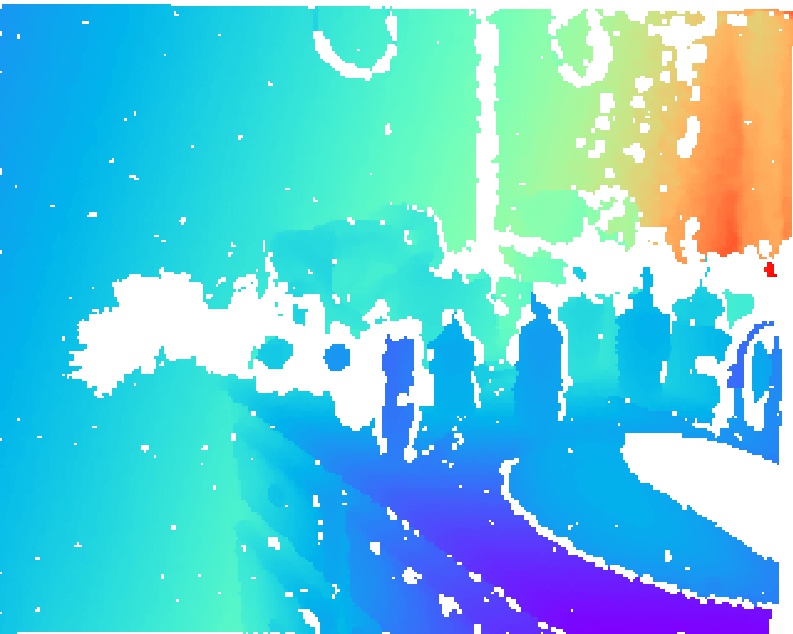}
        \caption{Quickshift \cite{quickshift}}
        \label{fig:quickshift}
      \end{subfigure}
      \begin{subfigure}[b]{0.32\linewidth}
        \centering
        \includegraphics[width=\linewidth]{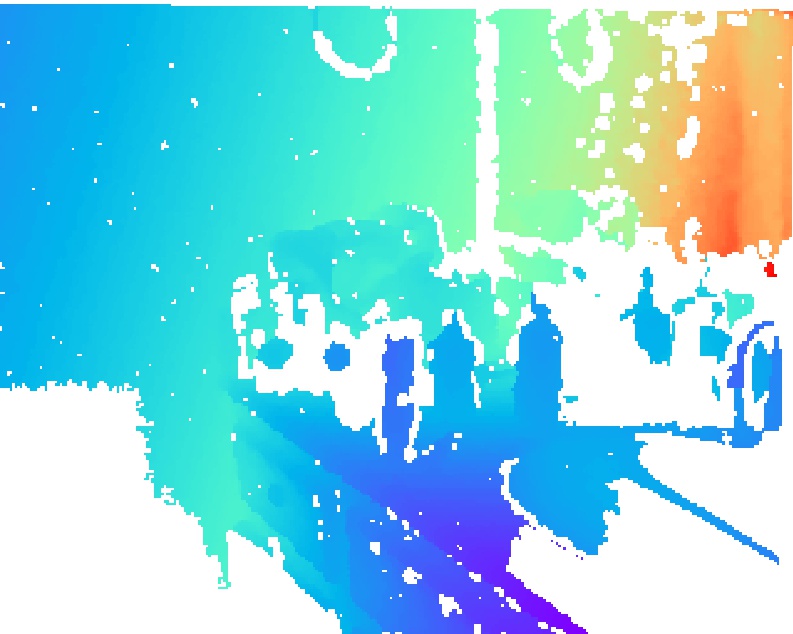}
        \caption{Slic \cite{slic}}
        \label{fig:slic}
        \centering
        \includegraphics[width=\linewidth]{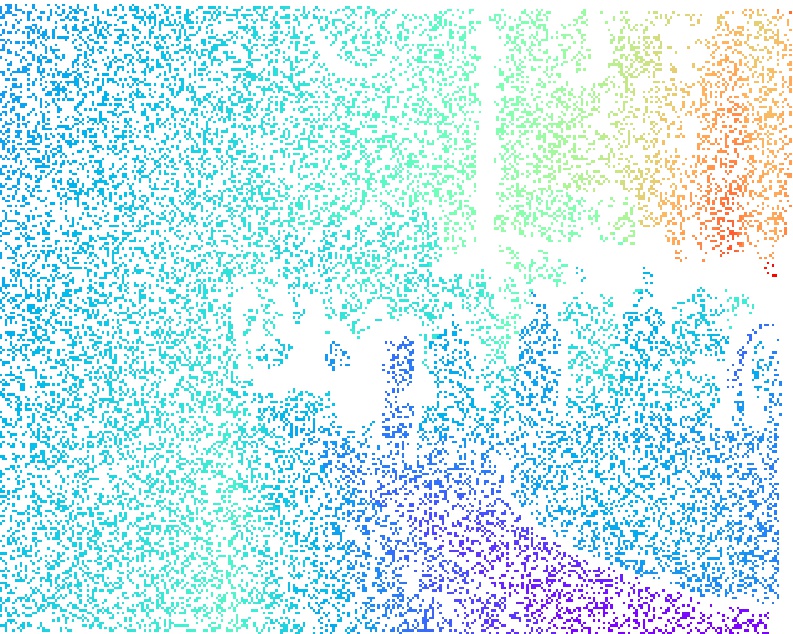}
        \caption{Uniform \cite{Ma2017SparseToDense}}
        \label{fig:lidar}
      \end{subfigure}
      \caption{Qualitative comparison of different sampling strategies based on classical image segmentation methods applied to a NYUv2~\cite{nyuv2} sample from the raw subset:
      \ref{fig:gt} and \ref{fig:rgb}~-- original image and depth map respectively;
      \ref{fig:felz} -- graph-based segmentation~\cite{felzenszwalb},
      \ref{fig:quickshift} -- quickshift segmentation (Vedaldi \etal \cite{quickshift}), \ref{fig:slic} -- SLIC segmentation (Achanta \etal \cite{slic}). All methods produce an image partition, then we replace depth data with zeros in segments with area below a predefined threshold value. Graph-based segmentation demonstrates the best match to the original sensor map, producing similar artifacts (\eg diffusion on the border of the table).}
      \label{fig:corrupt}
    \end{figure}

    Let $t \in T$ be a target sample that we want to degrade. Our goal is to construct a function $h: T \to S$ that transforms a depth map from the target domain $T$ to pseudo-sensor domain $S$. We assume that this procedure is sample-specific and can be factorized:
    $$
    h(\cdot) = z_g(\cdot | q) \circ z_n(\cdot) = z_n(z_g(\cdot | q)),
    $$
    where $q$ is the input RGB image. The term $z_g$ emulates a zero masking process guided by the image and $z_n$ is the zero masking caused by noise. The noise term $z_n$ represents a random spattering procedure uniformly distributed over the entire image. The specific form of $z_g$ may vary. Fig.~\ref{fig:corrupt} presents some possible approaches results. As shown in Figs.~\ref{fig:felz} and \ref{fig:gt}, the most suitable variant for semi-dense depth sensor simulation appears to be the graph-based segmentation algorithm introduced by Felzenszwalb and Huttenlocher~\cite{felzenszwalb}, thresholded by segment area. After obtaining pseudo-sensor data, we can perform a standard training procedure on it.
 
    Our corruption strategy (Fig.~\ref{fig:felz}) based on image segmentation significantly differs from widely-used sparse uniform sampling (Fig.~\ref{fig:lidar}). Below we compare these two strategies numerically on the NYUv2 dataset~\cite{nyuv2} using our model and additional open-source approaches from the KITTI dataset leaderboard~\cite{uhrig}. 
    
    \paragraph{Loss function.}

    Recent works underline two primary families of losses that are conceptually different: pixel-wise and pairwise. Pixel-wise loss functions measure the mean per-pixel distance between prediction and target, while their pairwise counterparts express the error by comparing the relationships between pairs of pixels $i$, $j$ in the output. The latter loss functions force the relationship between each pair of pixels in the prediction to be similar to that of the corresponding pair in the ground truth. In this work, we have experimented with several different single-term loss functions, including pair-wise and pixel-wise approaches in a logarithmic and actual domain (see supplementary materials for details). The logarithmic $L_1$ pair-wise loss function~\cite{romanov2020general} appears to be the most suitable for our network. It can be expressed as
    $$
            \mathcal{L}(y_i, y_i^*) = \frac{1}{|\mathcal{O}|^2} \sum_{i, j \in \mathcal{O}} \bigg| \log \frac{y_i}{y_j} - \log \frac{y_i^*}{y_j^*} \bigg|,
    $$
    where $\mathcal{O}$ is the set of pixels where the ground truth depth is non-zero, $i,j$ are pixel indices, $y_i, y^*_i$ are the predicted and target depth respectively. Following Eigen \etal~\cite{eigen_fergus}, our model predicts $\log y_i$ directly.
    
    \section{Experimental setup}\label{sec:experiments}


    \begin{figure*}[h]
    \setlength{\columnsep}{2pt}
    \setlength\multicolsep{0pt}
        \begin{multicols}{7}
        \includegraphics[width=1\linewidth]{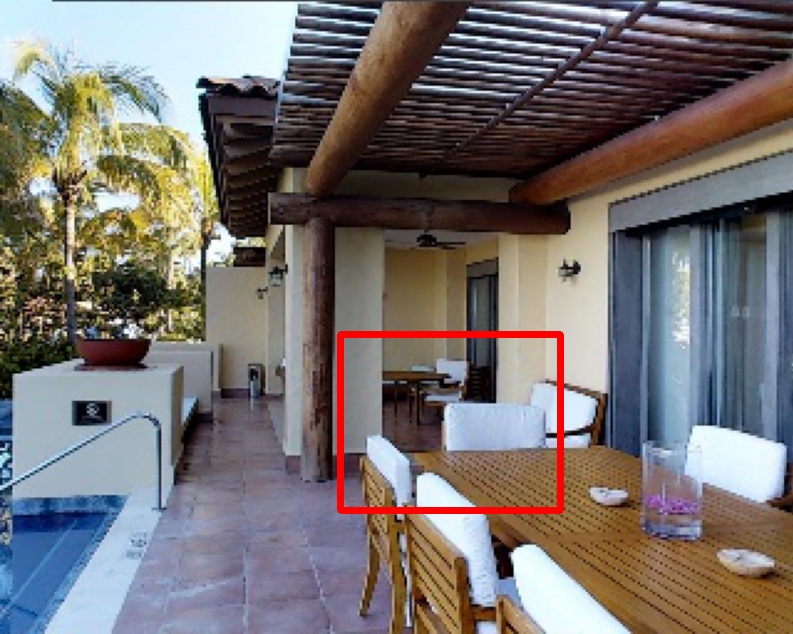}\par
        \includegraphics[width=1\linewidth]{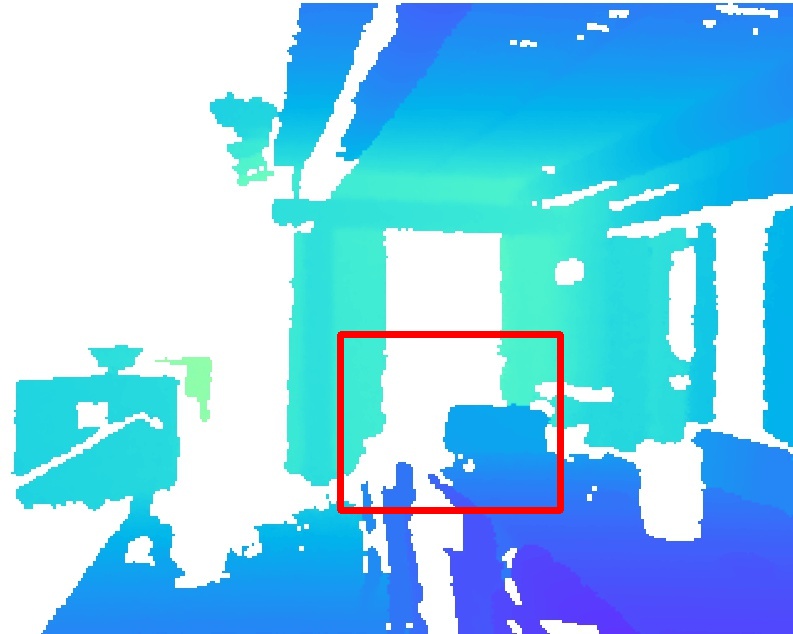}\par
        \includegraphics[width=1\linewidth]{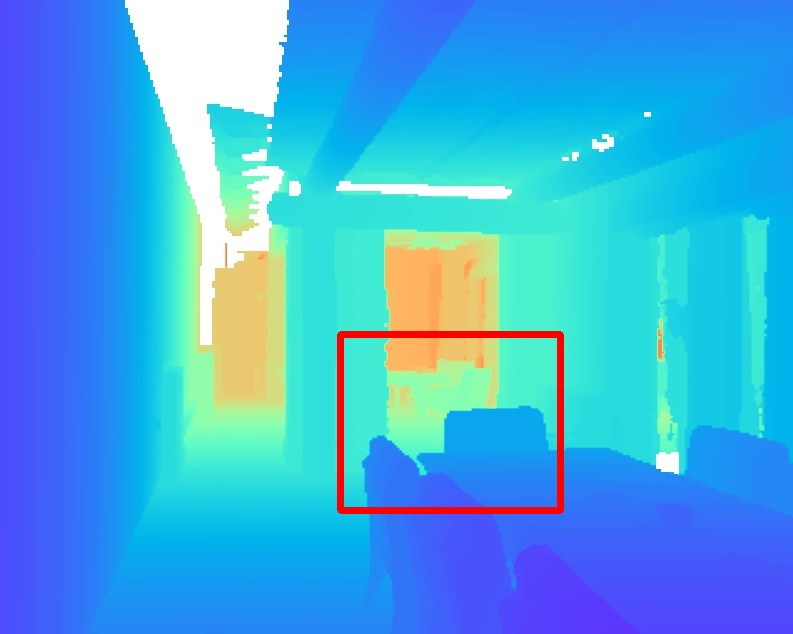}\par 
        \includegraphics[width=1\linewidth]{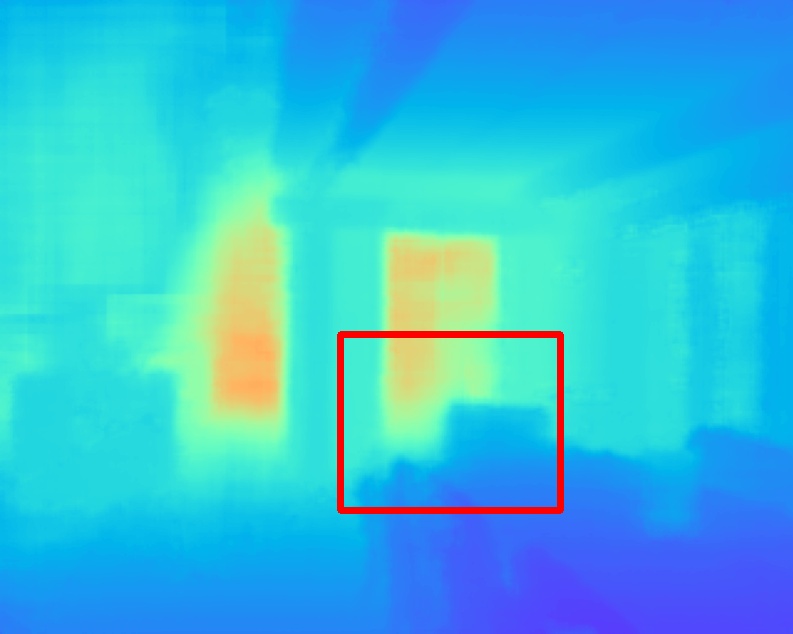}\par 
        \includegraphics[width=1\linewidth]{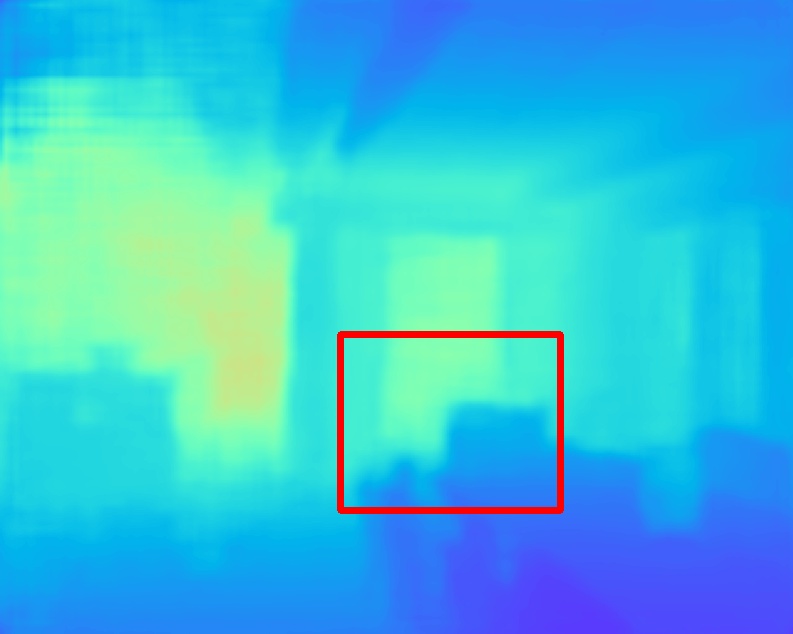}\par
        \includegraphics[width=1\linewidth]{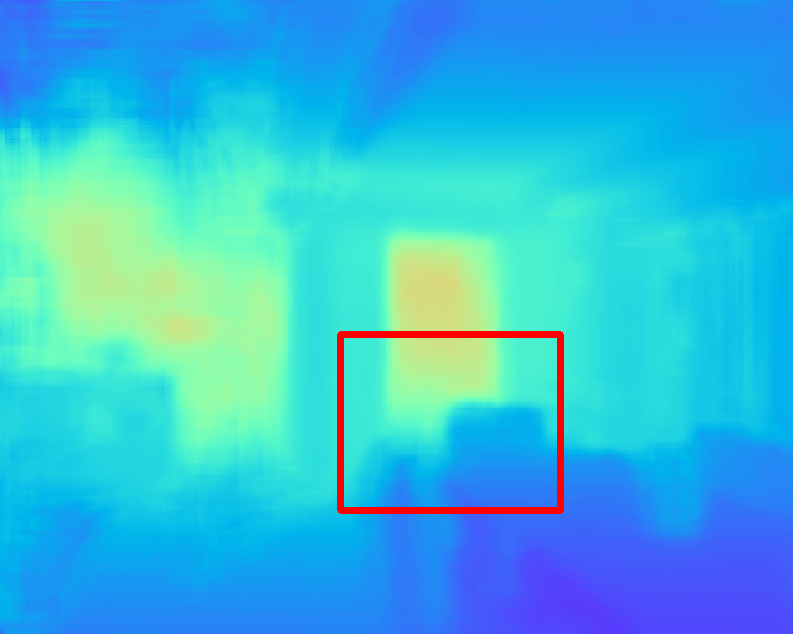}\par 
        \includegraphics[width=1\linewidth]{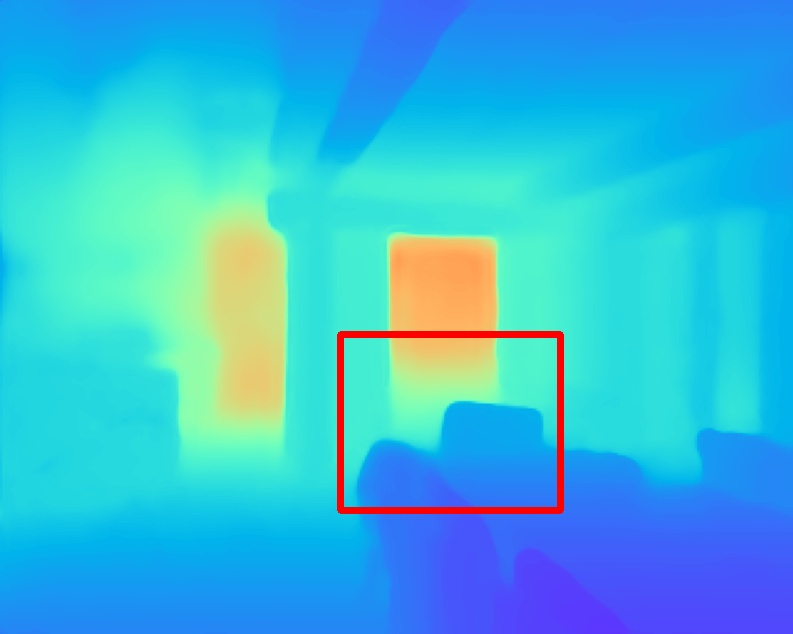}\par 
        \end{multicols}
        
        \begin{multicols}{7}
        \includegraphics[width=1\linewidth]{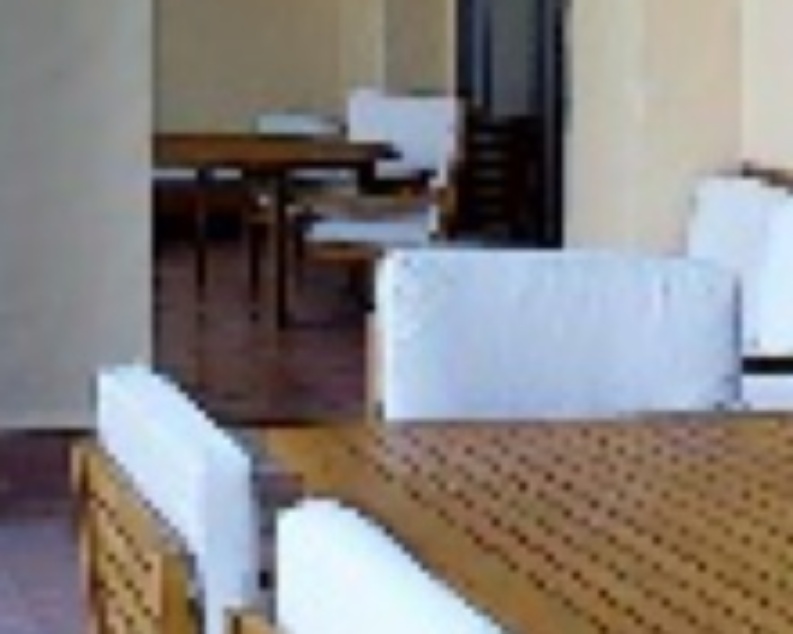}\par
        \includegraphics[width=1\linewidth]{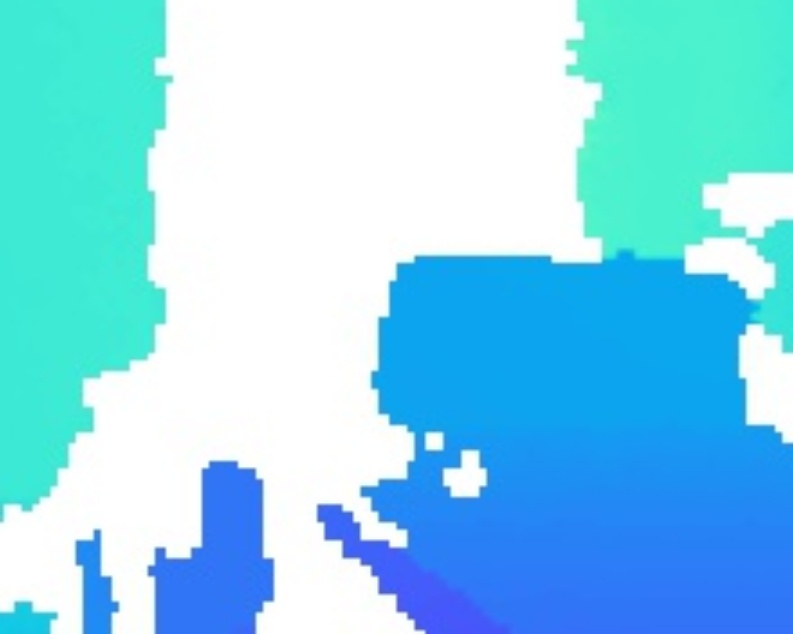}\par
        \includegraphics[width=1\linewidth]{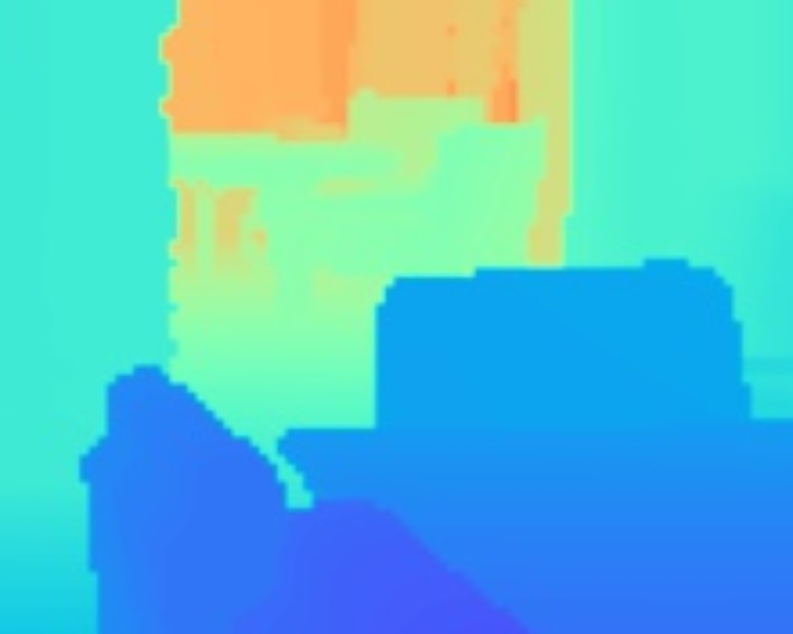}\par 
        \includegraphics[width=1\linewidth]{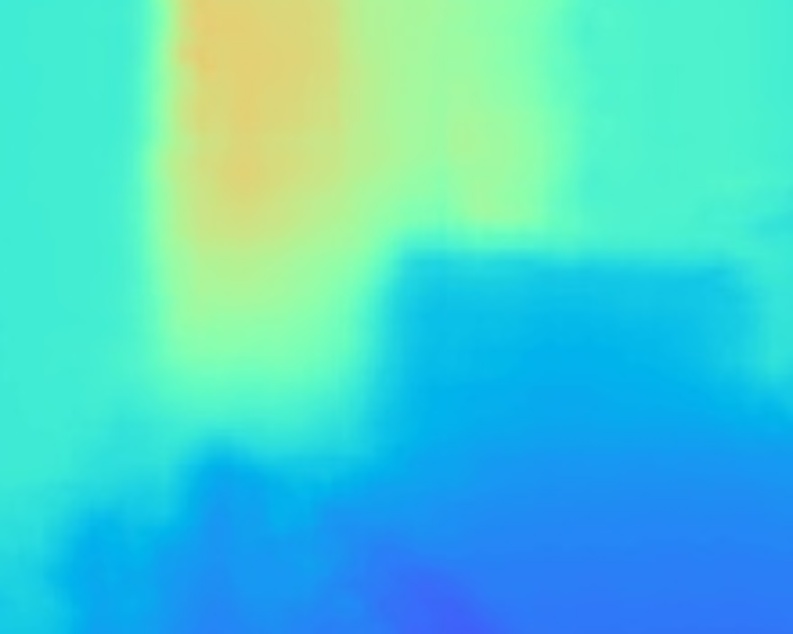}\par 
        \includegraphics[width=1\linewidth]{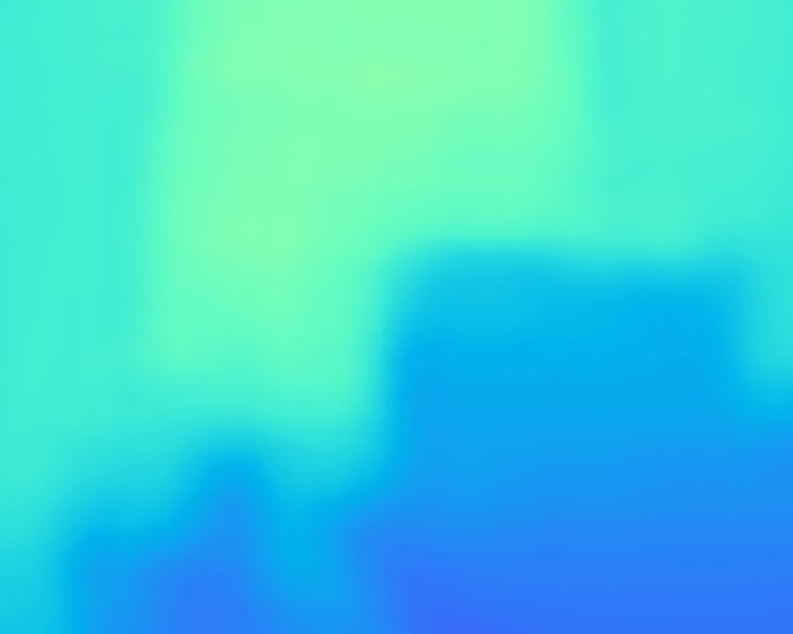}\par
        \includegraphics[width=1\linewidth]{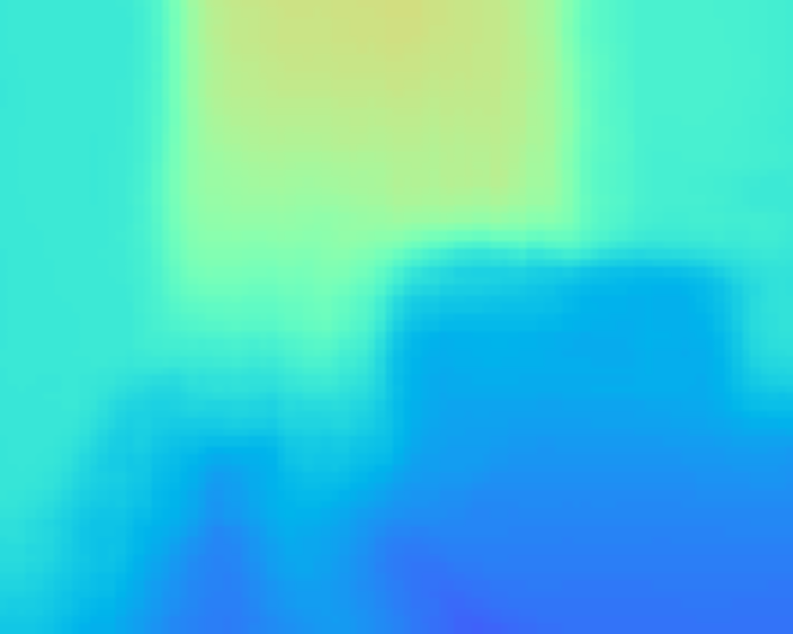}\par 
        \includegraphics[width=1\linewidth]{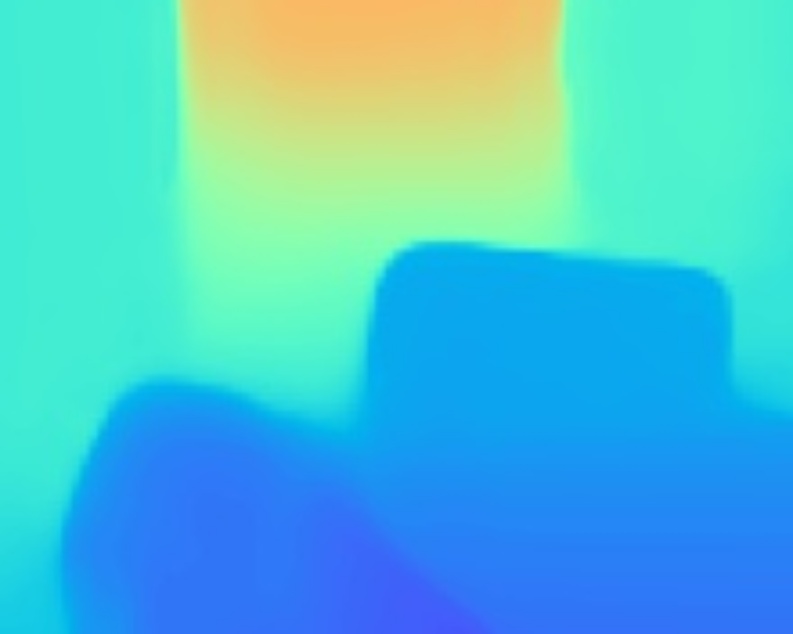}\par 
        \end{multicols}

        \begin{multicols}{7}
        \includegraphics[width=1\linewidth]{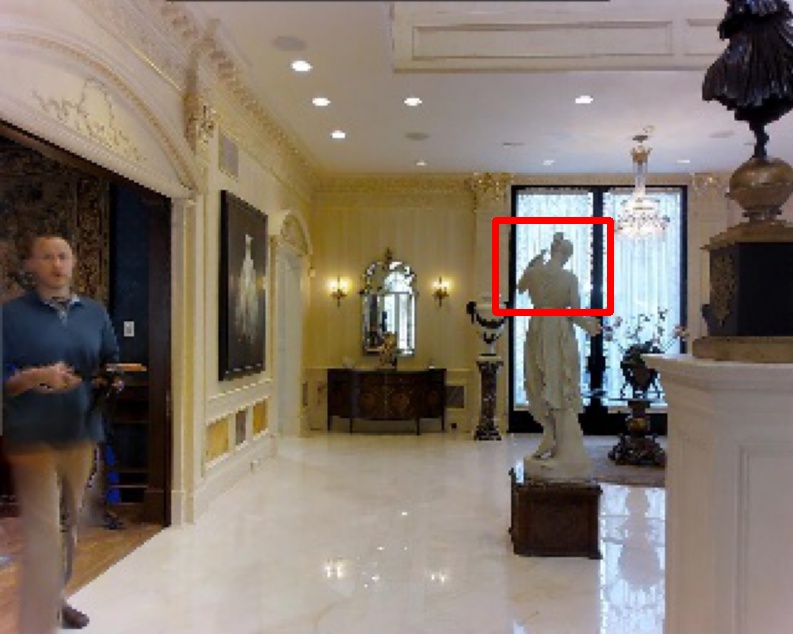}\par
        \includegraphics[width=1\linewidth]{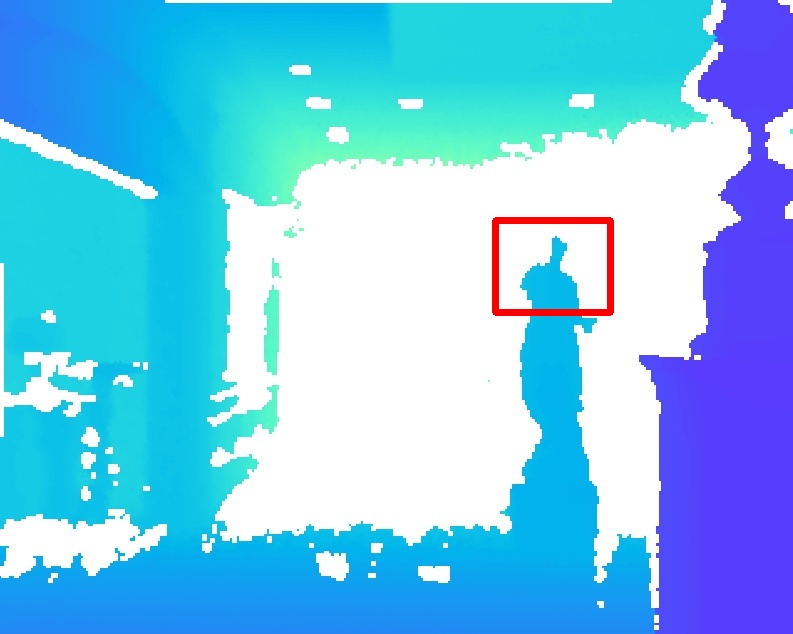}\par 
        \includegraphics[width=1\linewidth]{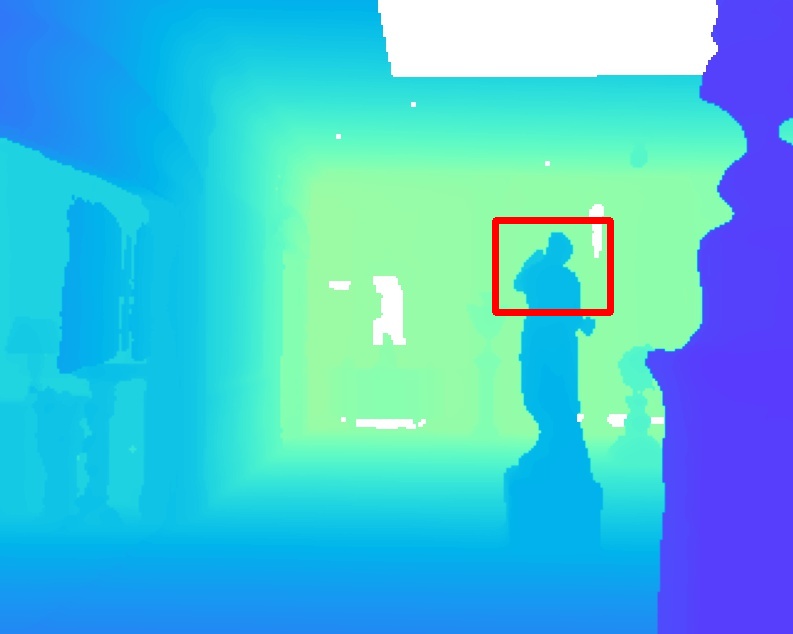}\par
        \includegraphics[width=1\linewidth]{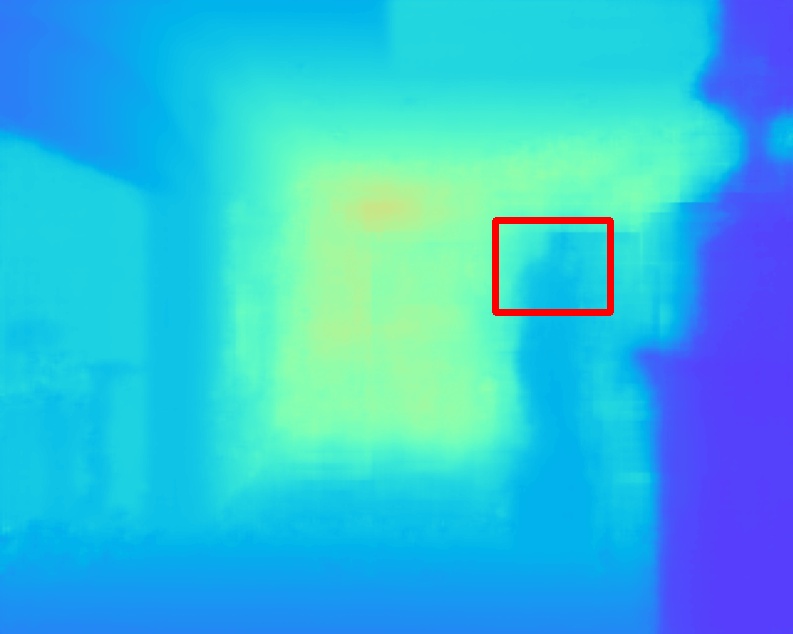}\par
        \includegraphics[width=1\linewidth]{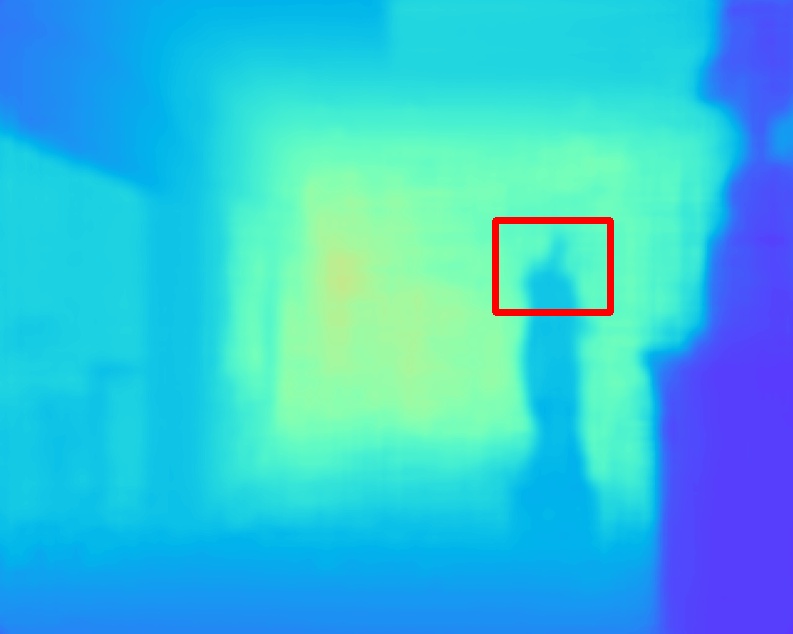}\par
        \includegraphics[width=1\linewidth]{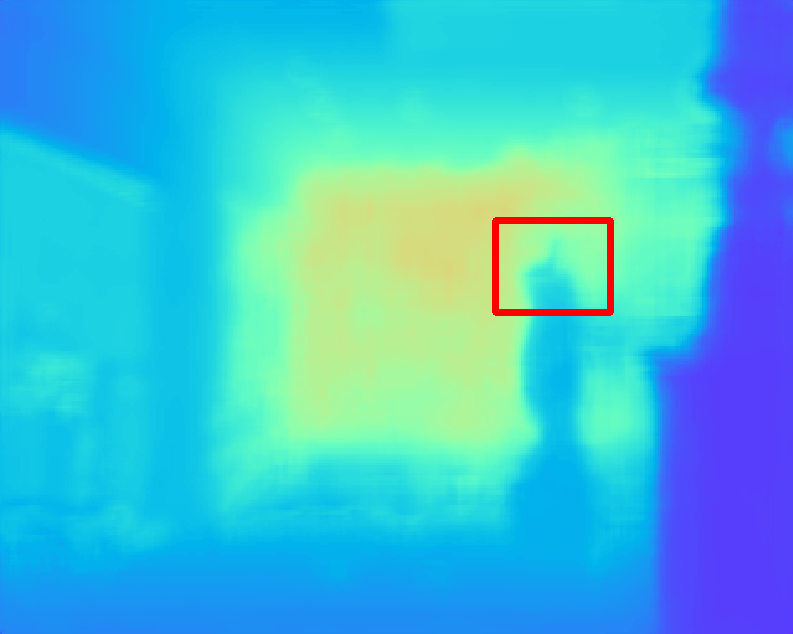}\par
        \includegraphics[width=1\linewidth]{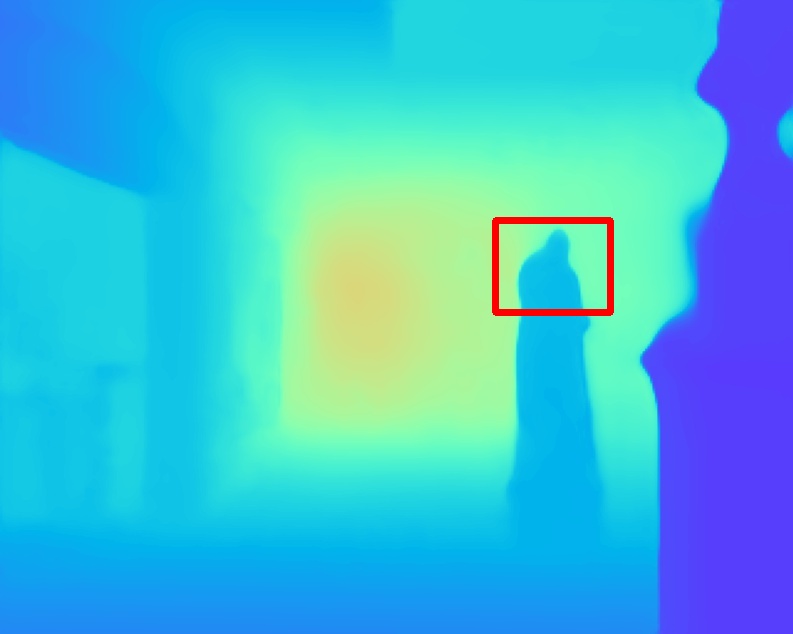}\par
        \end{multicols}
    
        \begin{multicols}{7}
        \includegraphics[width=1\linewidth]{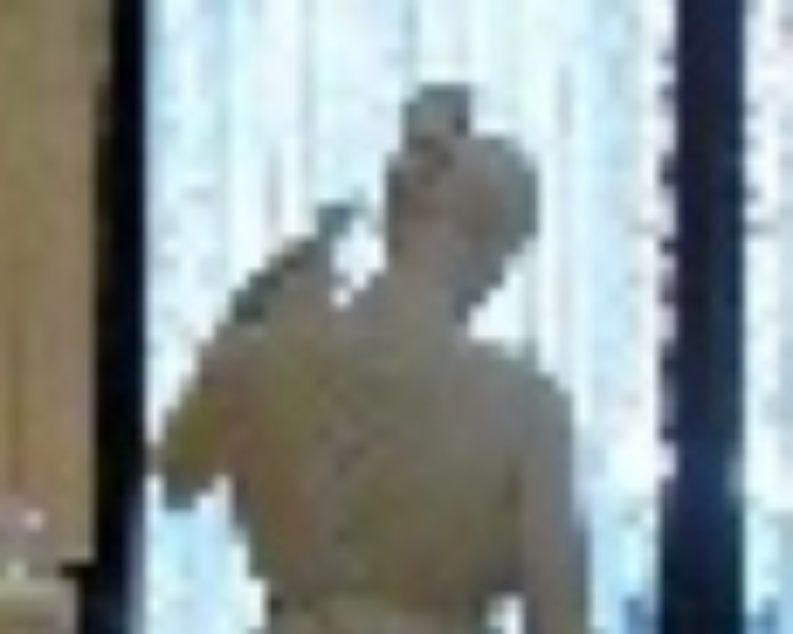} \centering \small{RGB}\par
        \includegraphics[width=1\linewidth]{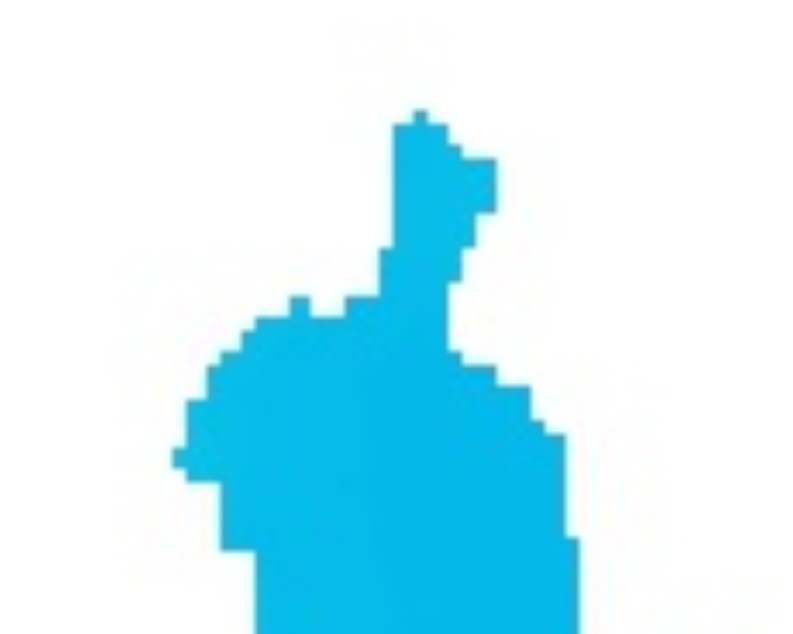} \centering \small{Sensor}\par 
        \includegraphics[width=1\linewidth]{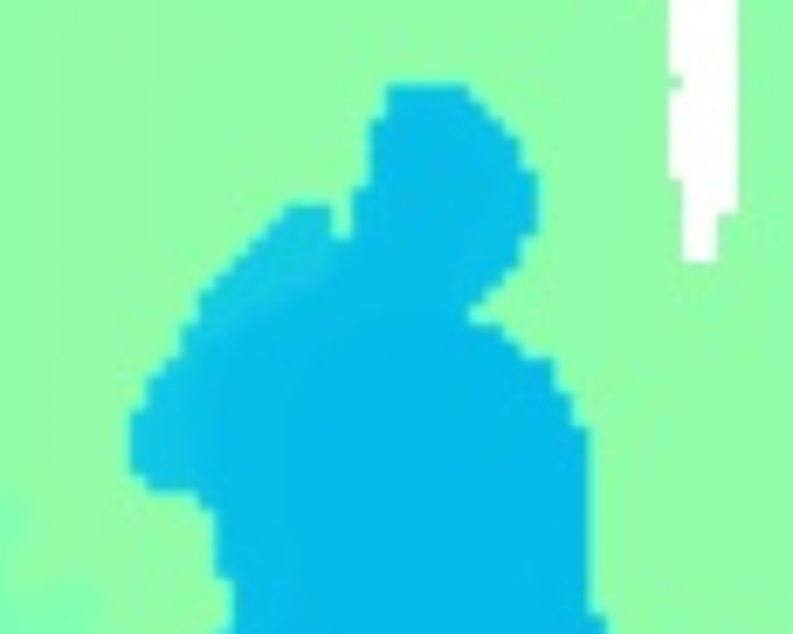} \centering \small{GT}\par
        \includegraphics[width=1\linewidth]{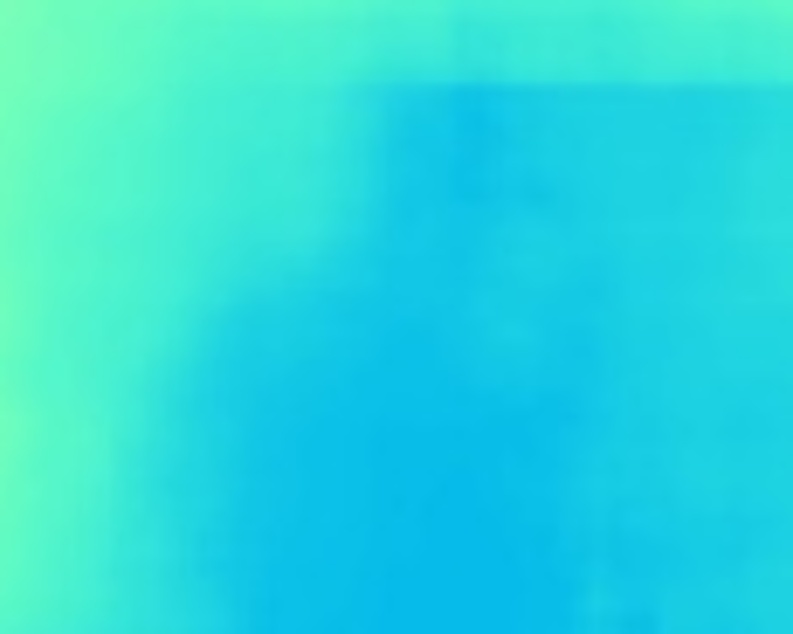}\centering \small{ Gansbeke \etal}\par
        \includegraphics[width=1\linewidth]{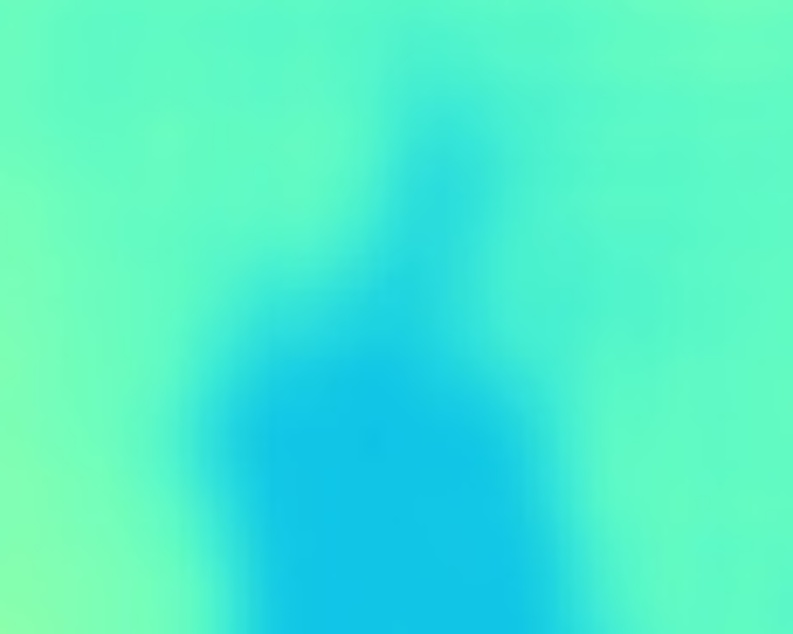} \centering \small{ Li \etal}\par
        \includegraphics[width=1\linewidth]{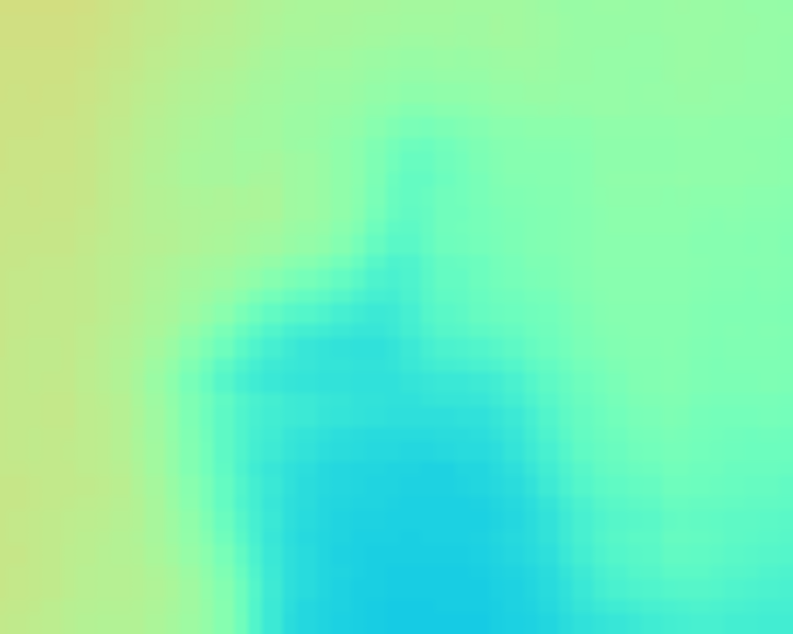} \centering \small{ Huang \etal}\par
        \includegraphics[width=1\linewidth]{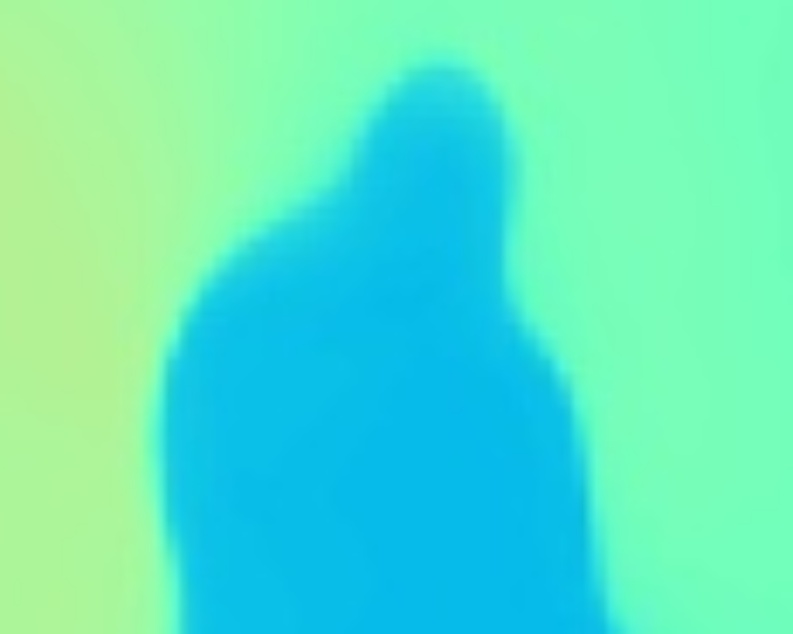} \centering \small{ Ours}\par
        \end{multicols}
    
    \caption{Qualitative comparison with Gansbeke \etal\cite{wvangansbeke_depth_2019}, Li \etal\cite{msg_chn}, Huang \etal\cite{Huang_2019} on Matterport3D test set. We train \cite{wvangansbeke_depth_2019} and \cite{msg_chn} on Matterport3D using the official code of the corresponding approaches, and results for \cite{Huang_2019} are based on the official pretrained model. Rows 2 and 4 represent zoomed-in fragments from rows 1 and 3, respectively. All images are created using color maps with the same value limits. Our model generates the completed depth map with very sharp boundaries.}
    \label{fig:mp3d_viz_test}
    \end{figure*}
    
    \begin{table*}[!th]
    \setlength{\tabcolsep}{8pt}
    \renewcommand{\arraystretch}{1.0}
    \centering
	\begin{tabular}{|c|c|c|c|c|c|c|c|c|}
	    \hline
	    & RMSE $\downarrow$ & MAE $\downarrow$ & $\delta_{1.05}$ $\uparrow$ & $\delta_{1.10}$ $\uparrow$ & $\delta_{1.25}$ $\uparrow$ & $\delta_{1.25^2}$ $\uparrow$ & $\delta_{1.25^3}$ $\uparrow$ & SSIM $\uparrow$ \\
		\hline
		\hline
		\textbf{Huang} \etal \cite{Huang_2019} & 1.092 & 0.342 & 0.661 & 0.750 & 0.850 & 0.911 & 0.936 & 0.799 \\
		\textbf{Zhang} \etal \cite{DBLP:journals/corr/abs-1803-09326} & 1.316 & 0.461 & 0.657 & 0.708 & 0.781 & 0.851 & 0.888 & 0.762 \\
		\textbf{Gansbeke} \etal \cite{wvangansbeke_depth_2019} & 1.161 & 0.395 & 0.542 & 0.657 & 0.799 & 0.887 & 0.927 & 0.700 \\
		\textbf{Li} \etal \cite{msg_chn} & 1.054 & 0.397 & 0.508 & 0.631 & 0.775 & 0.874 & 0.920 & 0.700 \\
		\makecell{\textbf{Gansbeke} \etal \cite{wvangansbeke_depth_2019} (ours)} & 1.264 & 0.484 & 0.675 & 0.741 & 0.826 & 0.888 & 0.920 & 0.780 \\
		\makecell{\textbf{Li} \etal \cite{msg_chn} (ours)} & 1.134 & 0.426 & 0.649 & 0.729 & 0.834 & 0.899 & 0.928 & 0.774 \\
		\hline
		\textbf{DM-LRN} (ours) & \textbf{0.961} & \textbf{0.285} & 0.726 & \textbf{0.813} & \textbf{0.890} & \textbf{0.933} & 0.949 & \textbf{0.844}\\
		\textbf{LRN} (ours) & 1.028 & 0.299 & 0.719 & 0.805 & \textbf{0.890} & 0.932 & \textbf{0.950} & 0.843 \\
		\textbf{LRN + mask} (ours) & 1.054 & 0.298 & \textbf{0.737} & \textbf{0.815} & 0.889 & \textbf{0.933} & \textbf{0.950} & \textbf{0.844} \\
		\hline
	\end{tabular}
	\vspace{0.1cm}
	\caption{\emph{Matterport3D TEST}. We use the results for Huang \etal \cite{Huang_2019} and Zhang \etal \cite{DBLP:journals/corr/abs-1803-09326} reported in~\cite{Huang_2019}. Gansbeke \etal \cite{wvangansbeke_depth_2019} and Li \etal \cite{msg_chn} are trained on Matterport3D using their official implementations. Models labeled as ``ours'' are trained using our proposed pipeline. The two bottom rows represent models without the decoder modulation branch, with and without the mask on the input. RMSE and MAE are measured in meters.}
	\label{tab:mp3d_test}
    \end{table*}
    
    \paragraph{Datasets.} 
    We perform comparative experiments on the following datasets: Matterport3D \cite{Matterport3D}, ScanNet \cite{dai2017scannet}, NYUv2 \cite{nyuv2} and KITTI\cite{uhrig}. Matterport3D includes real sensor data and ground truth depth data obtained from official reconstructed meshes. We use it as the primary target dataset. In order to investigate the generalization capabilities of the model, we perform validation of the models trained on the Matterport3D dataset directly on ScanNet. NYUv2 does not provide dense depth reconstruction for the entire dataset, so we evaluate our training strategy on this dataset. Although our approach is not intended to be applied to sparse depth sensors, we compare it with the best performing models on the KITTI dataset.

    \paragraph{Evaluation metrics}
    Following the standard evaluation protocol for indoor depth completion, we use root mean squared error (RMSE), mean absolute error (MAE), $\delta_i$, and SSIM. The $\delta_i$ metric denotes the percentage of predicted pixels where the relative error is less than a threshold $i$. Specifically, we evaluate $\delta_i$ for $i$ equal to $1.05$, $1.10$, $1.25$, $1.25^2$, and $1.25^3$; smaller values of $i$ correspond to making the $\delta_i$ metric more sensitive, while larger values reflect a more accurate prediction. RMSE and MAE directly measure absolute depth accuracy. RMSE is more sensitive to outliers than MAE and is usually chosen as the main metric for ranking models. In general, our testing pipeline for indoor depth completion is similar to Huang \etal~\cite{Huang_2019}.\footnote{The evaluation code is available on the official page \texttt{https://github.com/patrickwu2/Depth-Completion}. To keep a fair comparison, we opt for an evaluation procedure based on the official code.}. Following the KITTI leaderboard, we evaluate RMSE, MAE, iRMSE and iMAE metrics on the KITTI dataset.
    
    
    \begin{figure*}[!t]
    \setlength{\columnsep}{2pt}
    \setlength\multicolsep{0pt}
            \begin{multicols}{7}
            \includegraphics[width=1\linewidth]{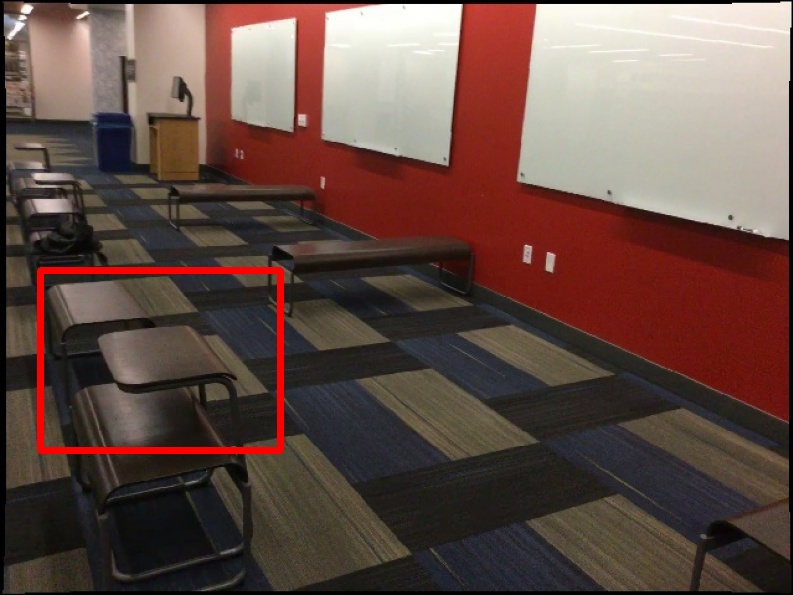}\par
            \includegraphics[width=1\linewidth]{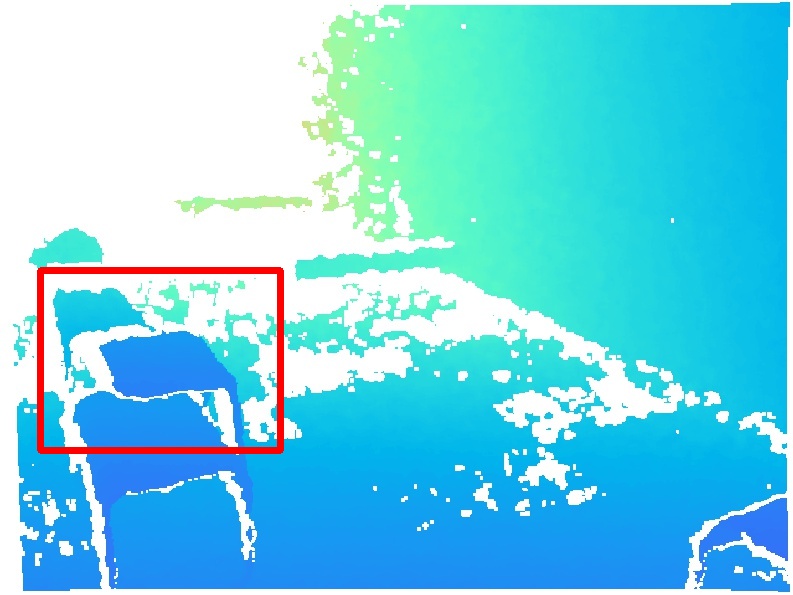}\par
            \includegraphics[width=1\linewidth]{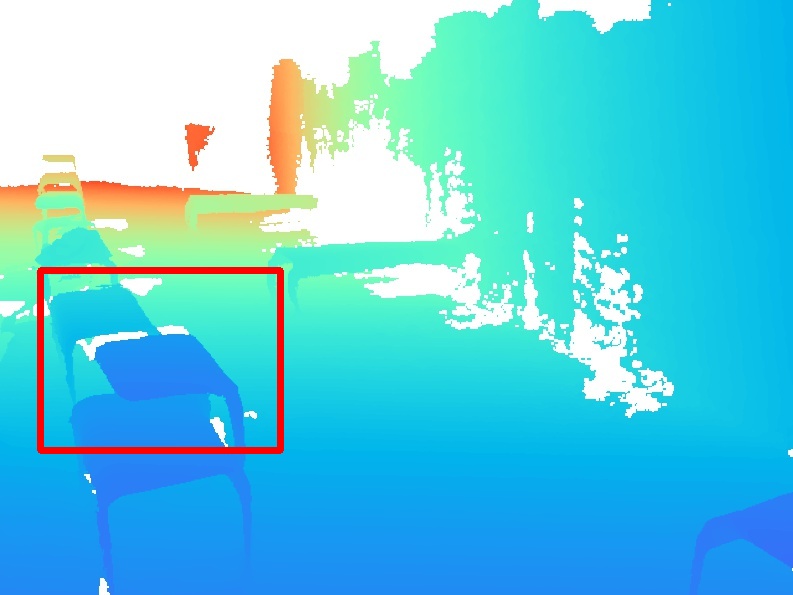}\par 
            \includegraphics[width=1\linewidth]{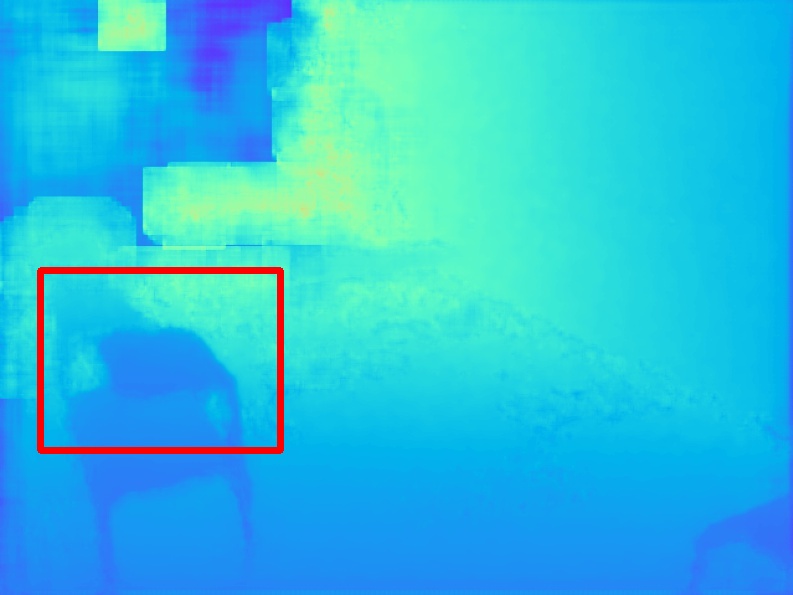}\par 
            \includegraphics[width=1\linewidth]{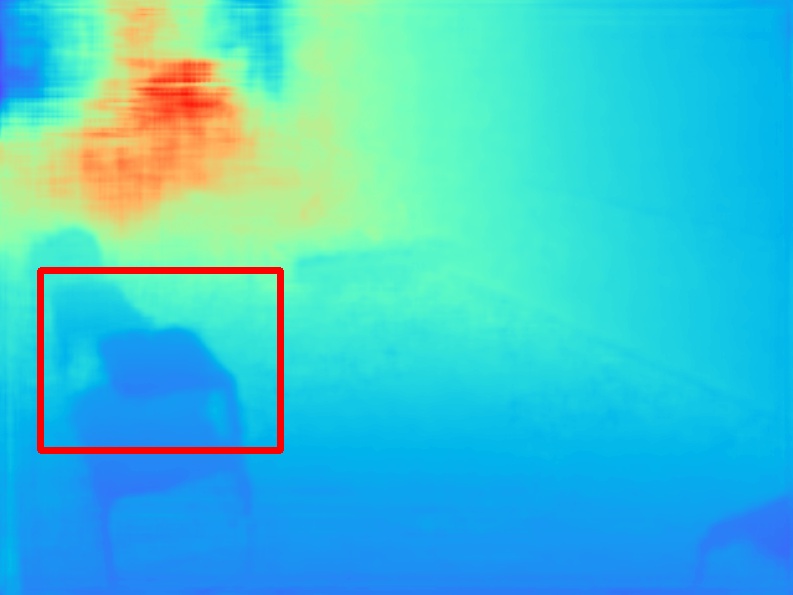}\par
            \includegraphics[width=1\linewidth]{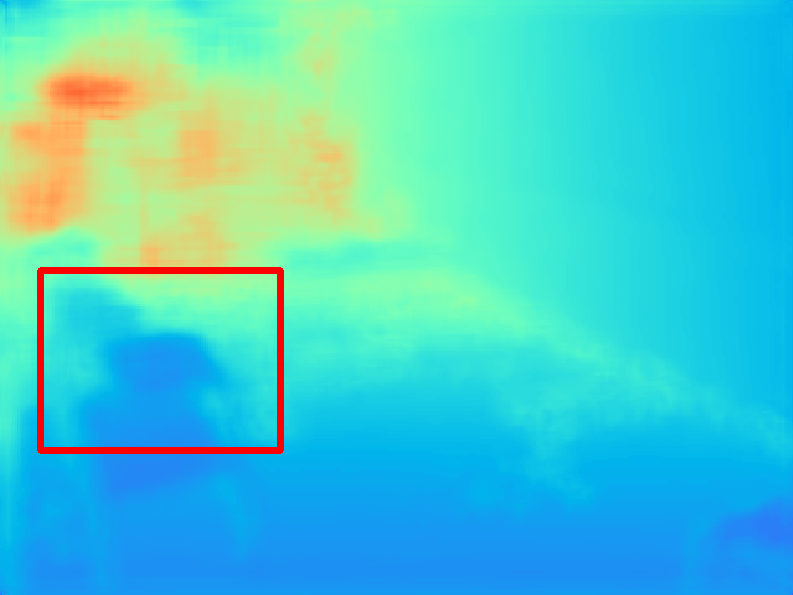}\par 
            \includegraphics[width=1\linewidth]{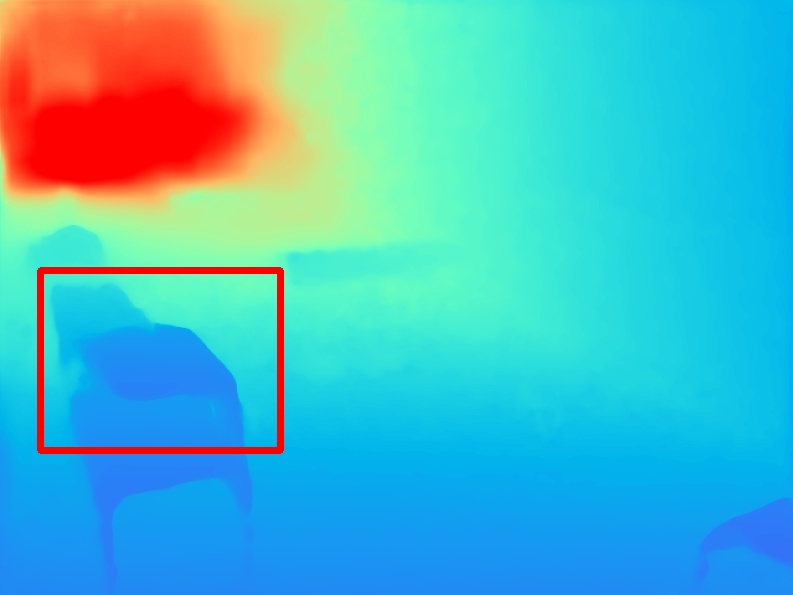}\par 
        \end{multicols}
        \begin{multicols}{7}
            \includegraphics[width=1\linewidth]{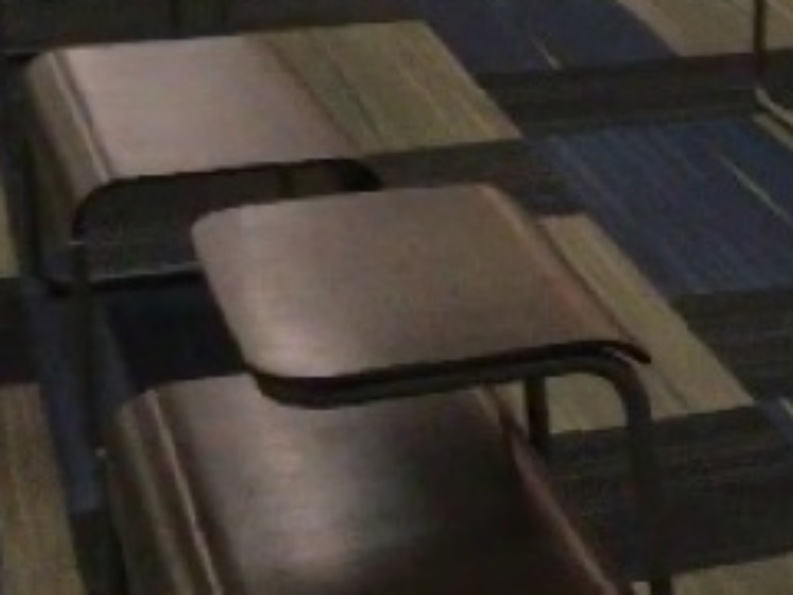}\par
            \includegraphics[width=1\linewidth]{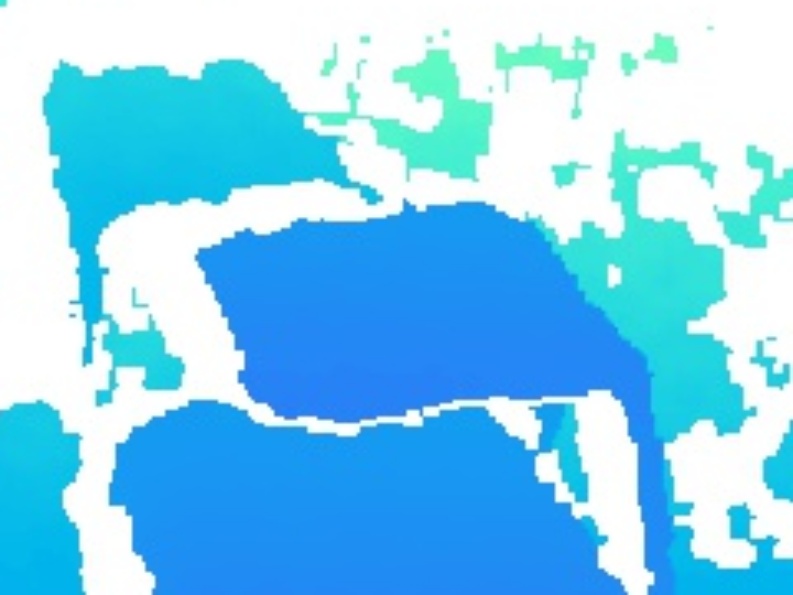}\par
            \includegraphics[width=1\linewidth]{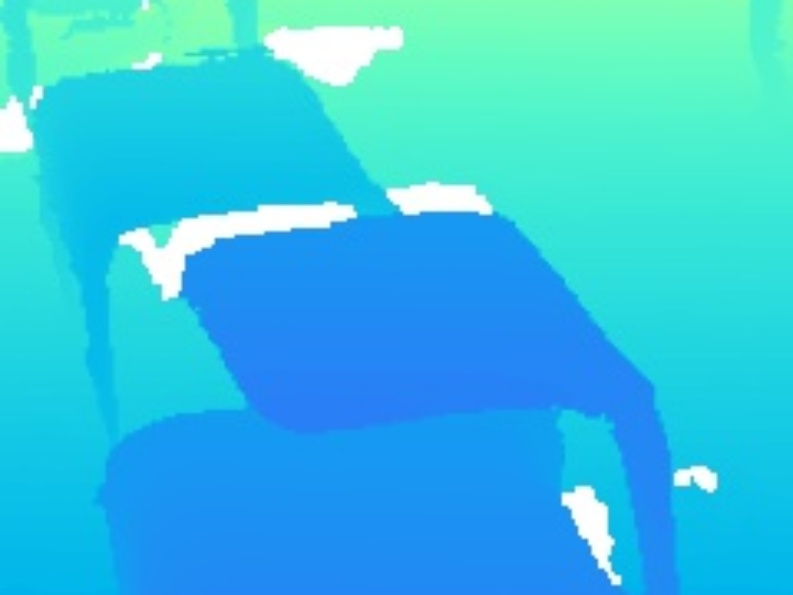}\par 
            \includegraphics[width=1\linewidth]{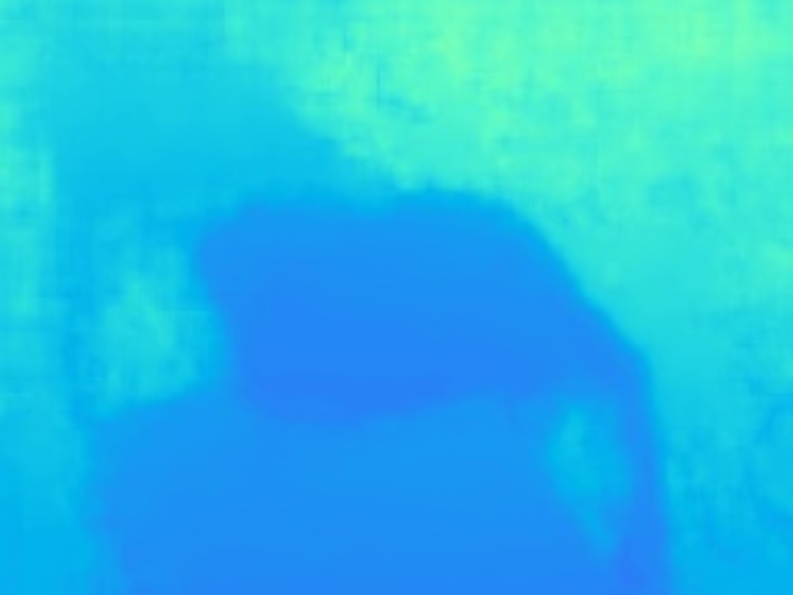}\par 
            \includegraphics[width=1\linewidth]{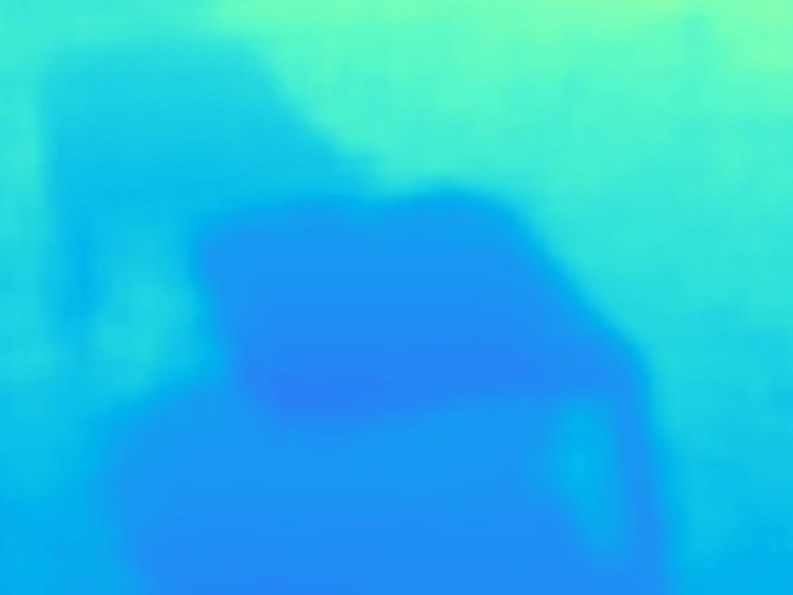}\par
            \includegraphics[width=1\linewidth]{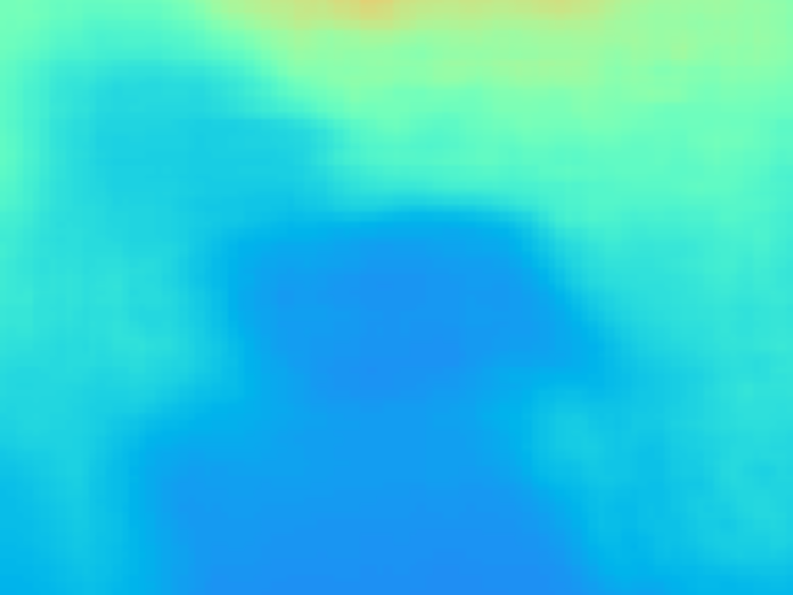}\par 
            \includegraphics[width=1\linewidth]{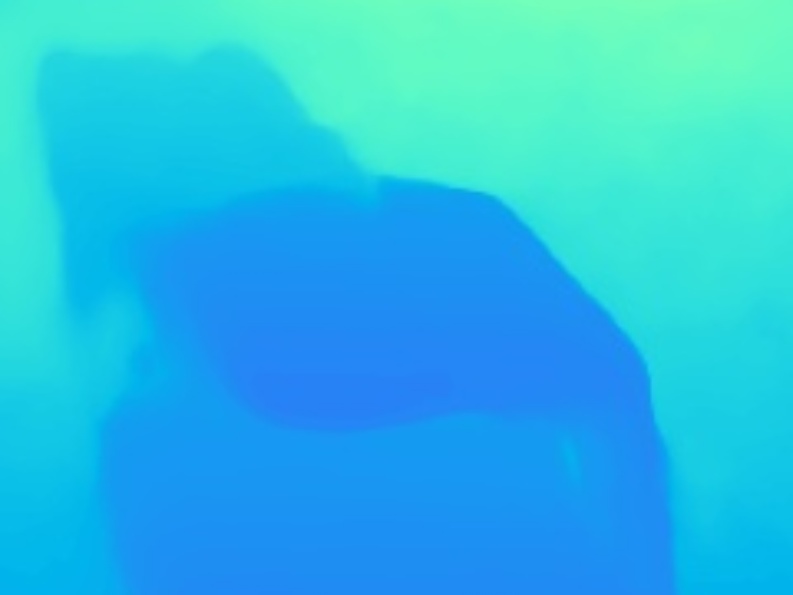}\par 
        \end{multicols}

        \begin{multicols}{7}
            \includegraphics[width=1\linewidth]{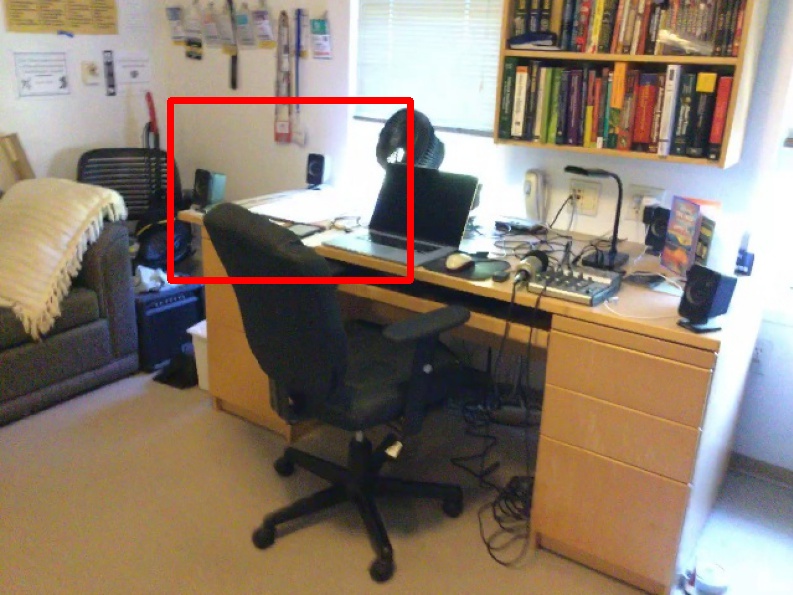}\par
            \includegraphics[width=1\linewidth]{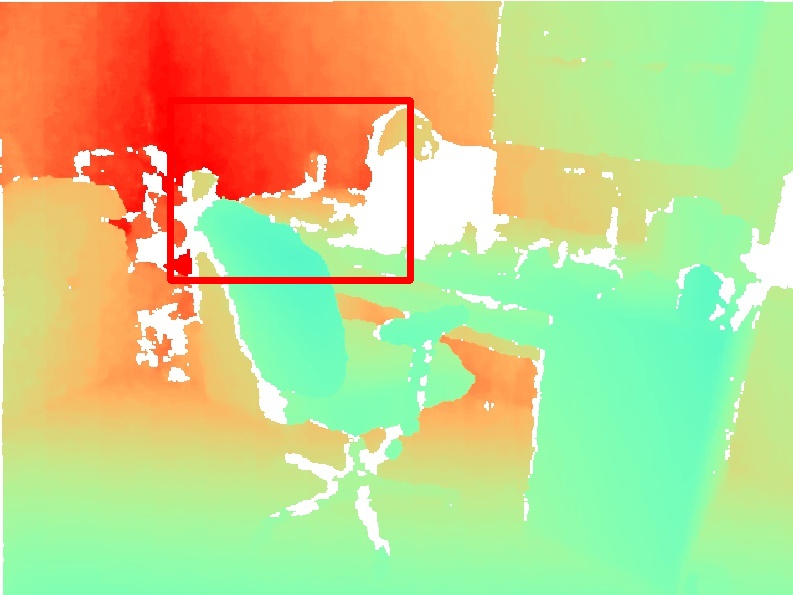}\par
            \includegraphics[width=1\linewidth]{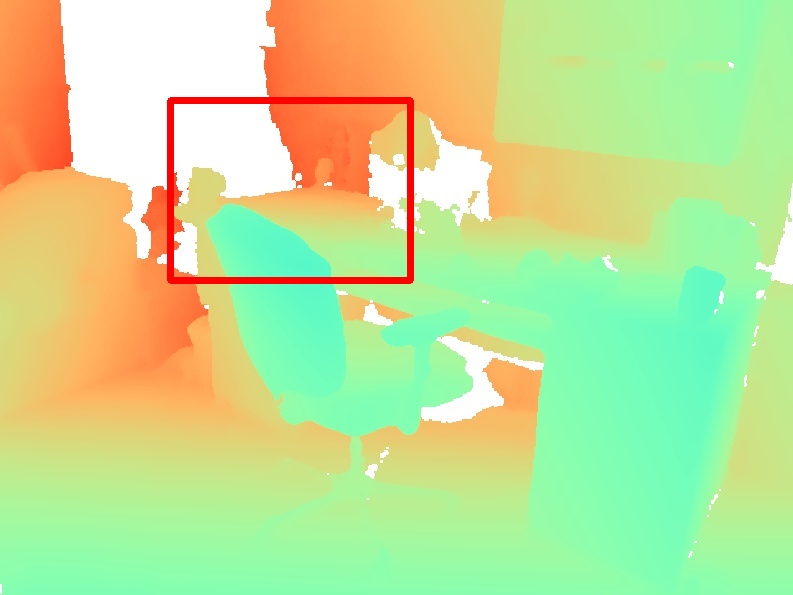}\par 
            \includegraphics[width=1\linewidth]{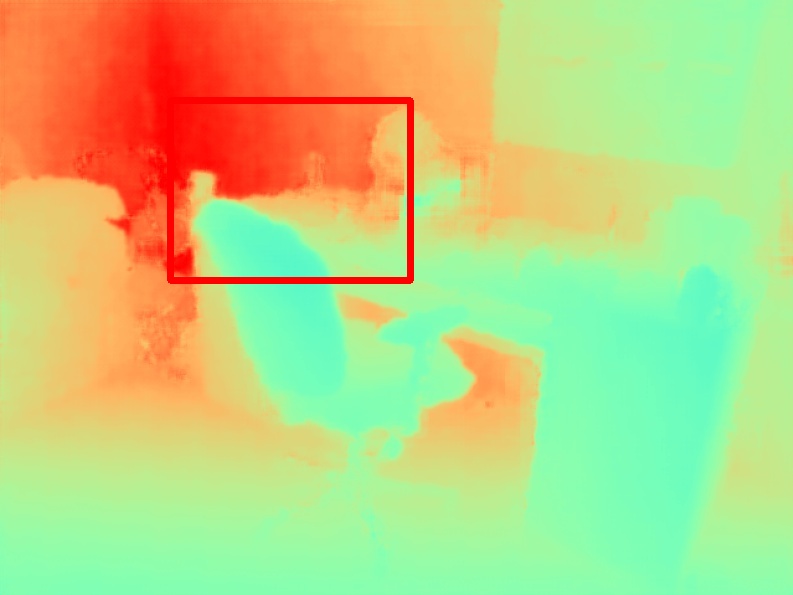}\par 
            \includegraphics[width=1\linewidth]{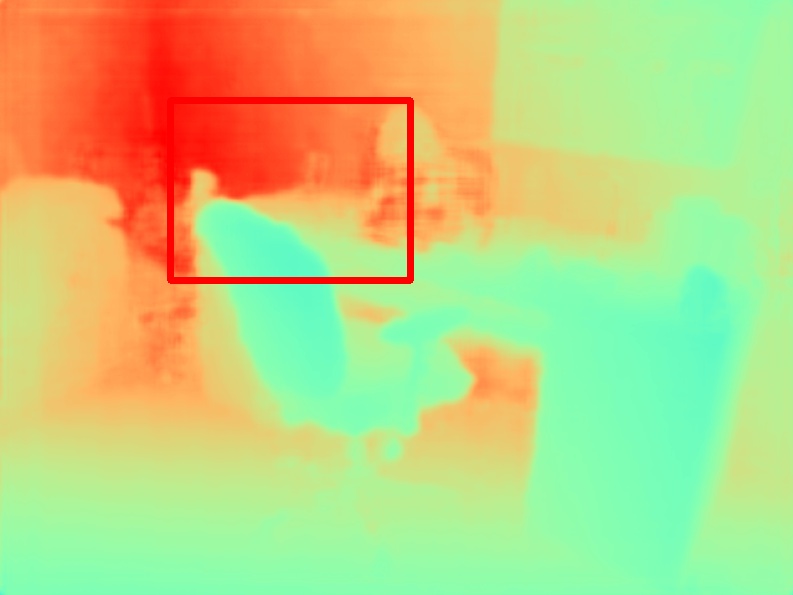}\par
            \includegraphics[width=1\linewidth]{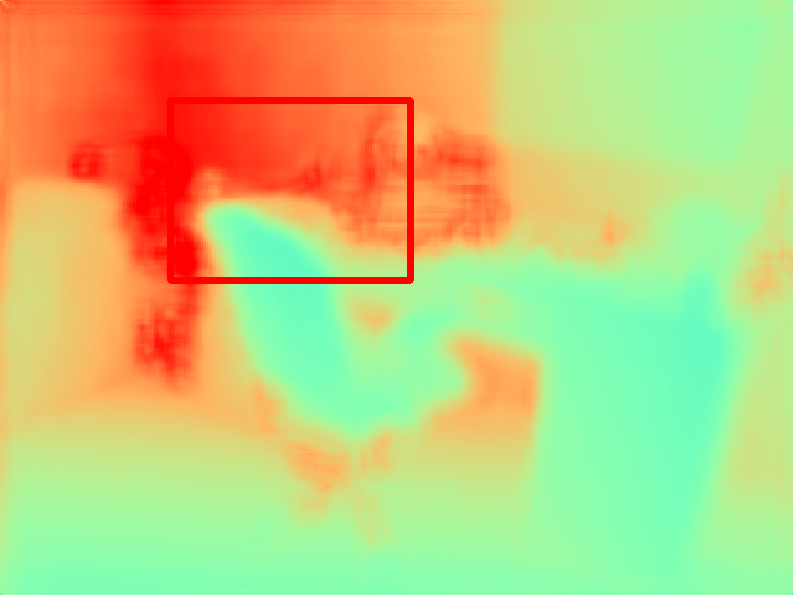}\par 
            \includegraphics[width=1\linewidth]{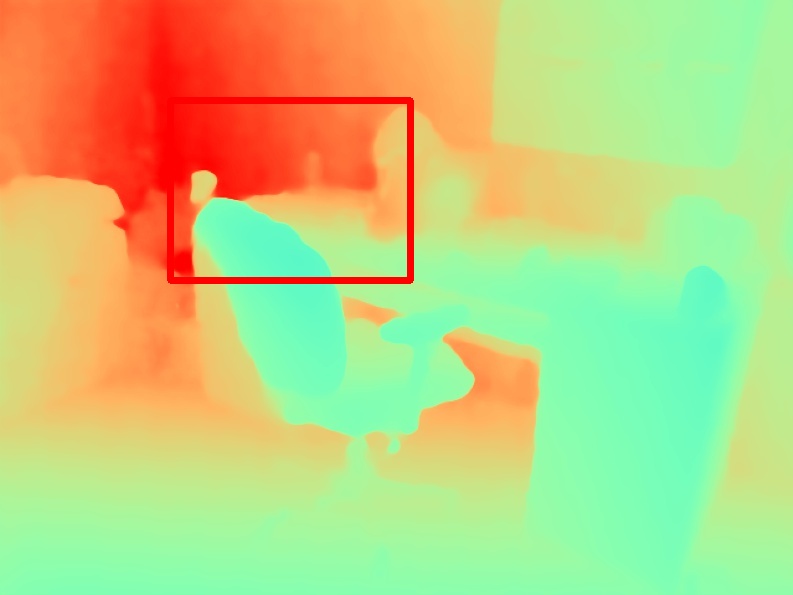}\par 
        \end{multicols}
        \begin{multicols}{7}
            \includegraphics[width=1\linewidth]{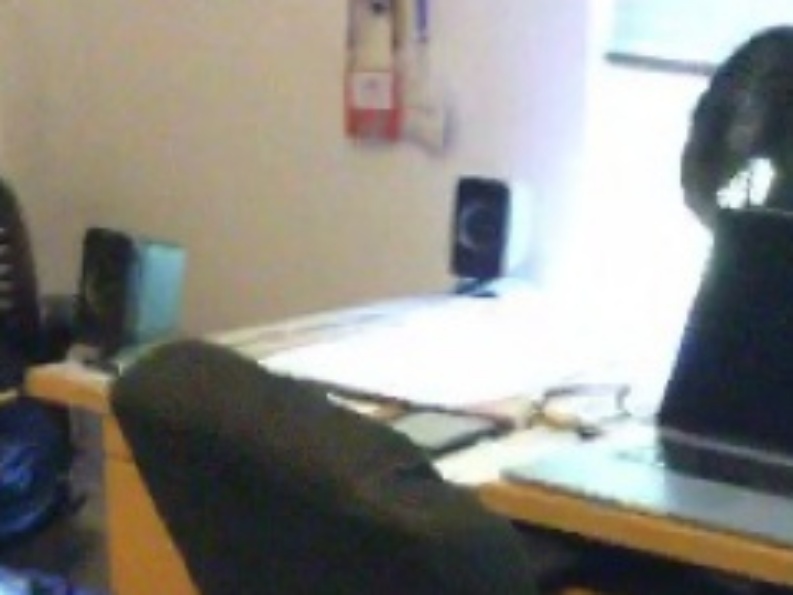} \centering \small{RGB}\par
            \includegraphics[width=1\linewidth]{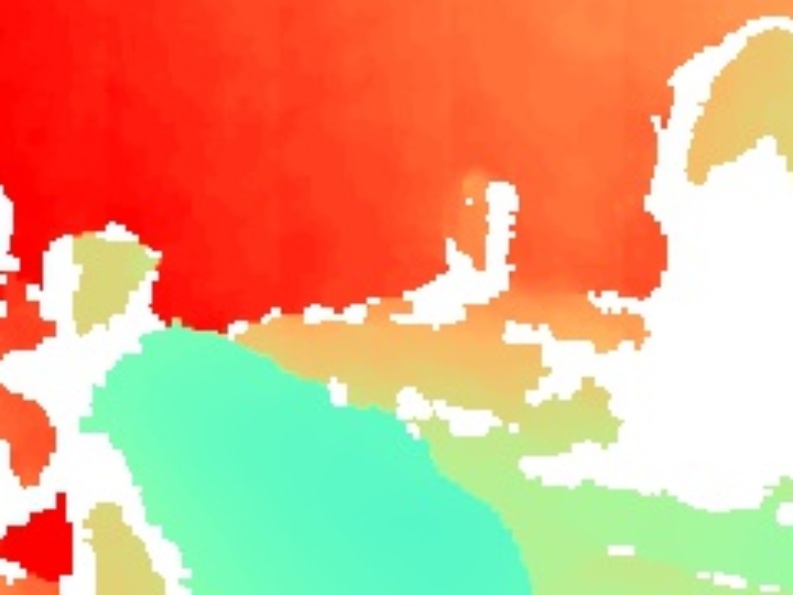} \centering \small{Sensor}\par
            \includegraphics[width=1\linewidth]{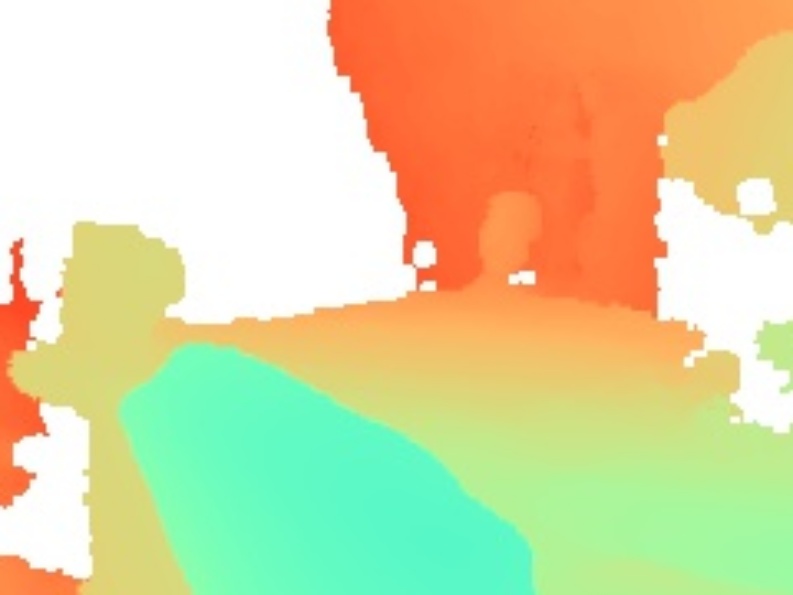} \centering \small{GT}\par 
            \includegraphics[width=1\linewidth]{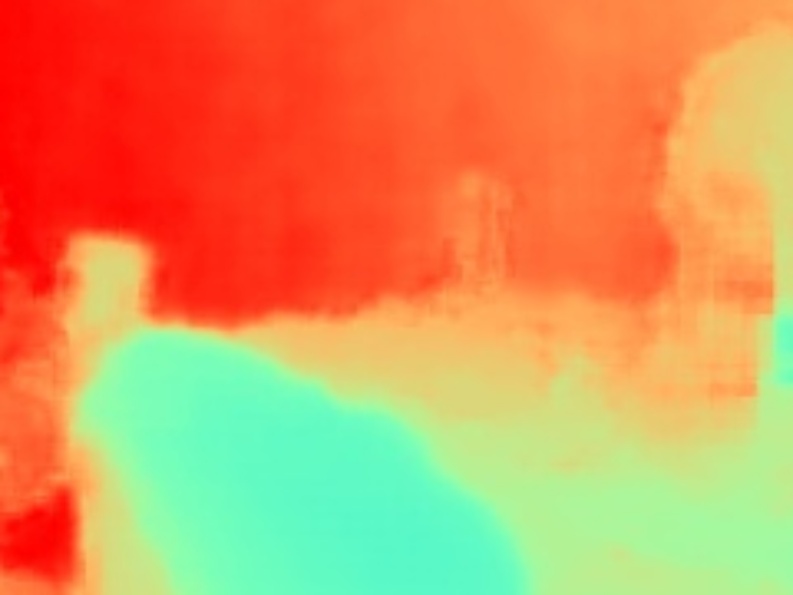}\centering \small{ Gansbeke \etal}\par 
            \includegraphics[width=1\linewidth]{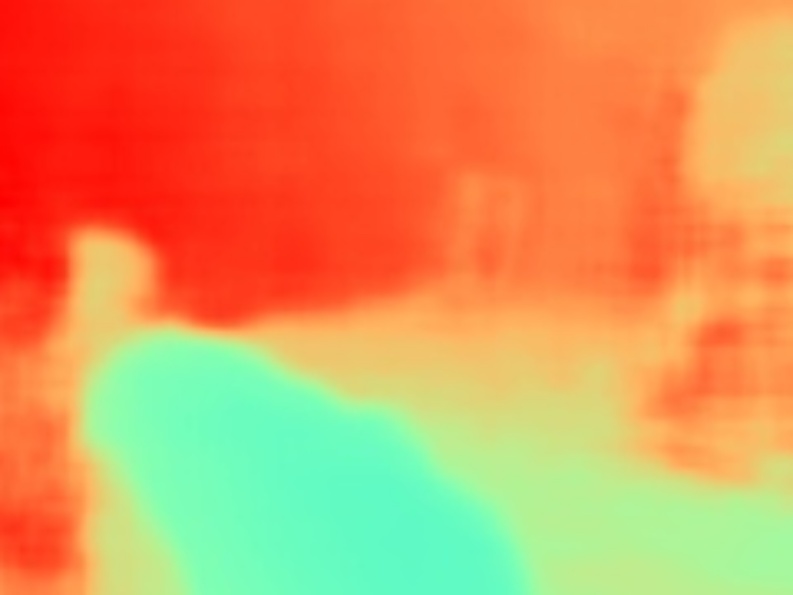} \centering \small{ Li \etal}\par
            \includegraphics[width=1\linewidth]{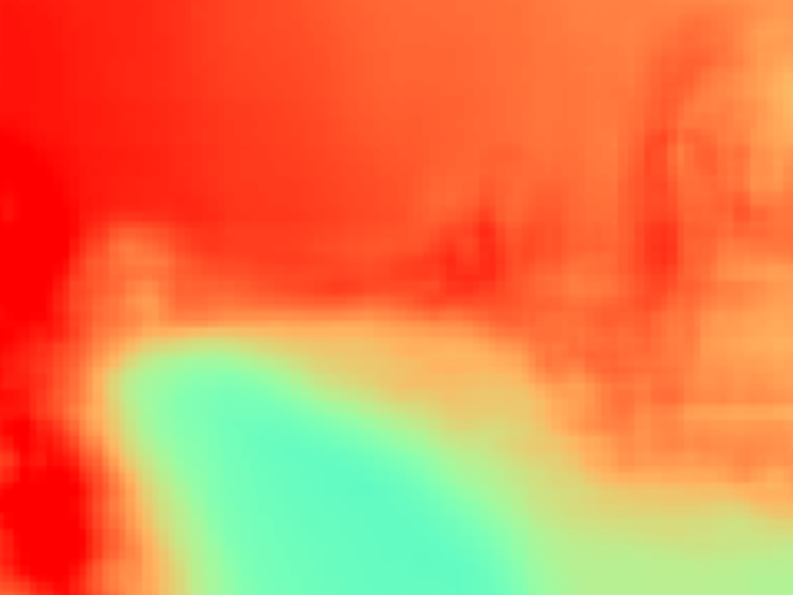} \centering \small{ Huang \etal}\par 
            \includegraphics[width=1\linewidth]{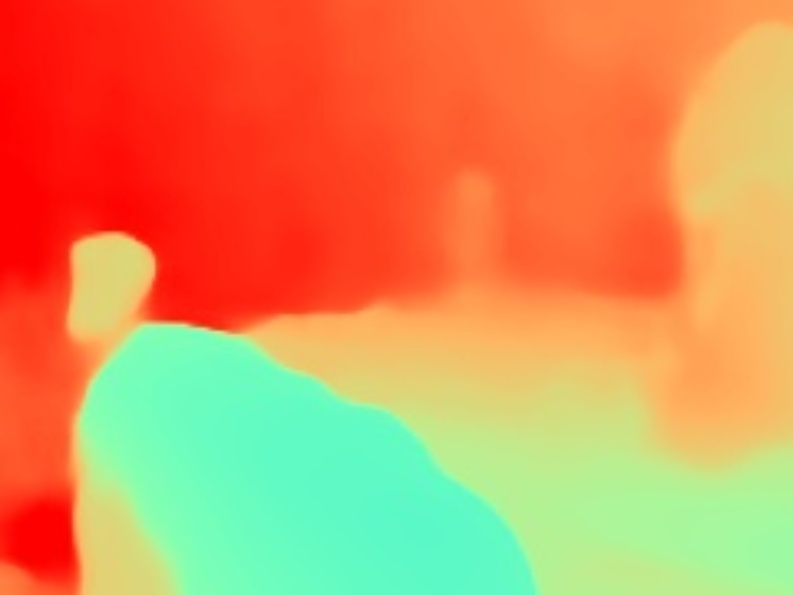} \centering \small{ Ours}\par 
        \end{multicols}

    \caption{Qualitative comparison with Gansbeke \etal\cite{wvangansbeke_depth_2019}, Li \etal\cite{msg_chn}, Huang \etal\cite{Huang_2019} on ScanNet \cite{dai2017scannet}. All models are trained on Matterport3D. The images are received using a unified color map.}
    \label{fig:scannet_viz_test}
    \end{figure*}
    
    \begin{table*}[ht]
    \setlength{\tabcolsep}{9pt}
    \renewcommand{\arraystretch}{1.0}
    \centering
	\begin{tabular}{|c|c|c|c|c|c|c|c|c|}
	    \hline
	    & RMSE $\downarrow$ & MAE $\downarrow$ & $\delta_{1.05}$ $\uparrow$ & $\delta_{1.10}$ $\uparrow$ & $\delta_{1.25}$ $\uparrow$ & $\delta_{1.25^2}$ $\uparrow$ & $\delta_{1.25^3}$ $\uparrow$ & SSIM $\uparrow$ \\
		\hline
		\hline
		\textbf{Huang \etal \cite{Huang_2019}} & 0.244 & 0.097 & 0.736 & 0.850 & 0.945 & 0.982 & 0.992 & 0.812 \\
		\textbf{Zhang \etal \cite{DBLP:journals/corr/abs-1803-09326}} & 0.214 & 0.080 & 0.769 & 0.881 & 0.958 & 0.985 & 0.993 & 0.850 \\
		\textbf{Gansbeke} \etal \cite{wvangansbeke_depth_2019} & 0.223 & 0.074 & 0.829 & 0.899 & 0.954 & 0.980 & 0.990 & 0.850 \\
		\textbf{Li} \etal \cite{msg_chn} & \textbf{0.190} & 0.067 & 0.828 & 0.903 & 0.961 & \textbf{0.986} & \textbf{0.995} & 0.875 \\
		\hline
		\textbf{DM-LRN} (ours) & 0.198 & \textbf{0.054} & \textbf{0.900} & \textbf{0.933} & \textbf{0.962} & 0.982 & 0.992 & \textbf{0.918}\\
		\hline
	\end{tabular}
	\vspace{0.1cm}
	\caption{\emph{ScanNet TEST}. Cross-dataset testing demonstrates the strong generalization capability of our method. All models are trained on Matterport3D. RMSE and MAE are measured in meters.}
	\label{tab:scannet_test}
    \end{table*}

    \paragraph{Implementation details}
    In our experiments, we use the Adam \cite{adam} optimizer with initial learning rate set to $10^{-4}$, $\beta_1 = 0.9$, $\beta_2 = 0.999$ and without weight decay. The pretrained EfficientNet-b4 \cite{EffitientNet} backbone is used unless otherwise stated. Batch normalization is controlled by the modulation process, so we fine-tune its parameters during the first epoch only, and afterwards these parameters are fixed. The training process is performed end-to-end for 100 epochs on a single Nvidia Tesla P40 GPU. We implement all models in Python 3.7 using the PyTorch library~\cite{NEURIPS2019_9015}.

\section{Results}\label{sec:Matterport3D}

    \paragraph{Matterport3D.} We begin by inferencing our indoor pipeline on the \emph{Matterport3D} dataset. Since very few previous approaches have been tested and achieved good results on this dataset, we train some of the best performing open-source KITTI models~\cite{msg_chn, wvangansbeke_depth_2019} for a fair comparison. Assuming that the original training pipeline of these models might be designed specifically for LiDAR data, we also perform a complementary training procedure in our training setup. 
    
    The results of this quantitative comparison are presented in Table~\ref{tab:mp3d_test}. 
    Our training pipeline applied to KITTI models improves the results in terms of $\delta_i$, especially with smaller values of $i$, but leads to artifacts captured by RMSE values. The original training setup of these methods also does not show state of the art performance on \emph{Matterport3D} (see Table~\ref{tab:mp3d_test}). We use the original training procedure for further experiments. These methods do not produce sharp edges (see Fig.~\ref{fig:mp3d_viz_test}) that are crucial for indoor applications. Zhang \etal~\cite{DBLP:journals/corr/abs-1803-09326} and Huang \etal~\cite{Huang_2019} managed to address this problem and received less blurry results. Our model produces improved completed depth while being more accurate in terms of both RMSE and MAE. In Table~\ref{tab:mp3d_test}, we also present ablation experiments including different masking strategies.  
    
    A visual comparison is shown in Figure~\ref{fig:mp3d_viz_test}. Our model keeps the sensor data almost unchanged and sharp. 
    Moreover, the geometric shapes of the interior layout and objects in the scene remain distinct.
    

    \begin{figure*}[!t]
    \setlength{\columnsep}{2pt}
    \setlength\multicolsep{0pt}
            \begin{multicols}{7}
            \includegraphics[width=1\linewidth]{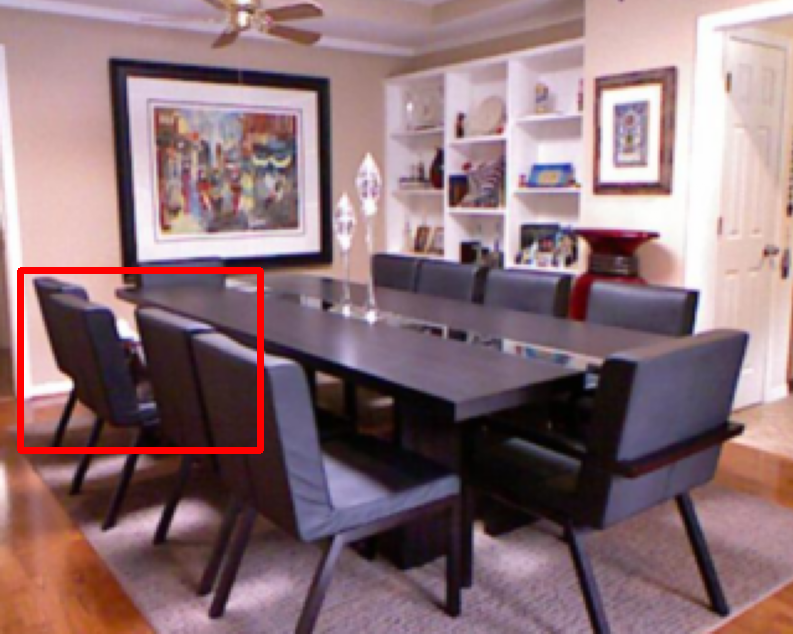}\par
            \includegraphics[width=1\linewidth]{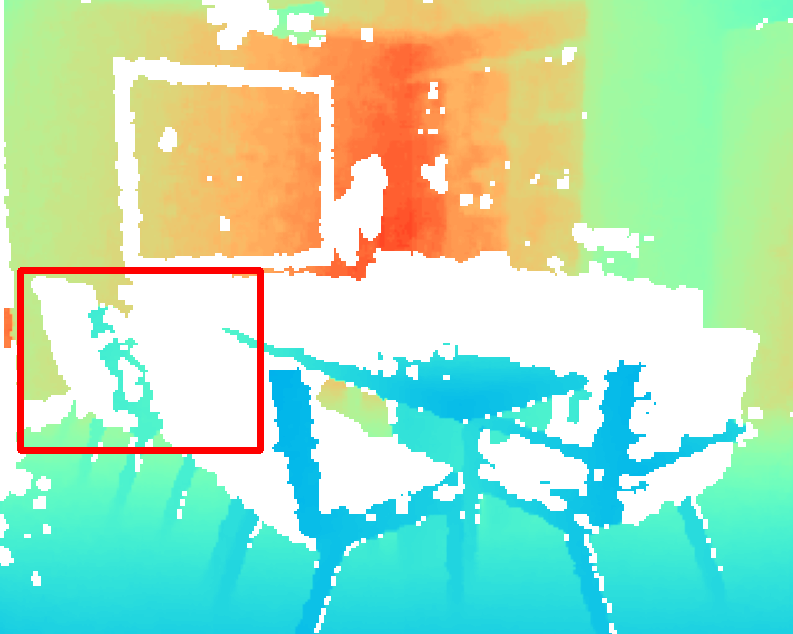}\par
            \includegraphics[width=1\linewidth]{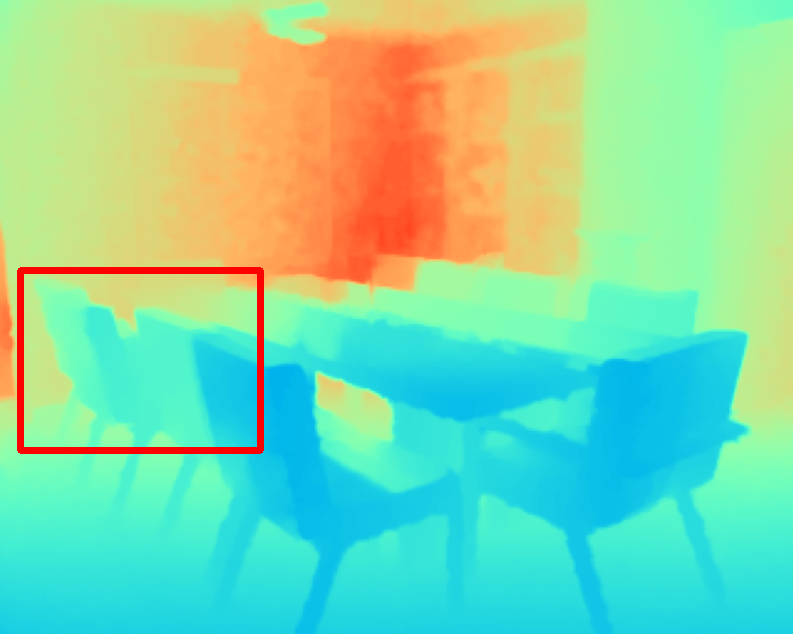}\par 
            \includegraphics[width=1\linewidth]{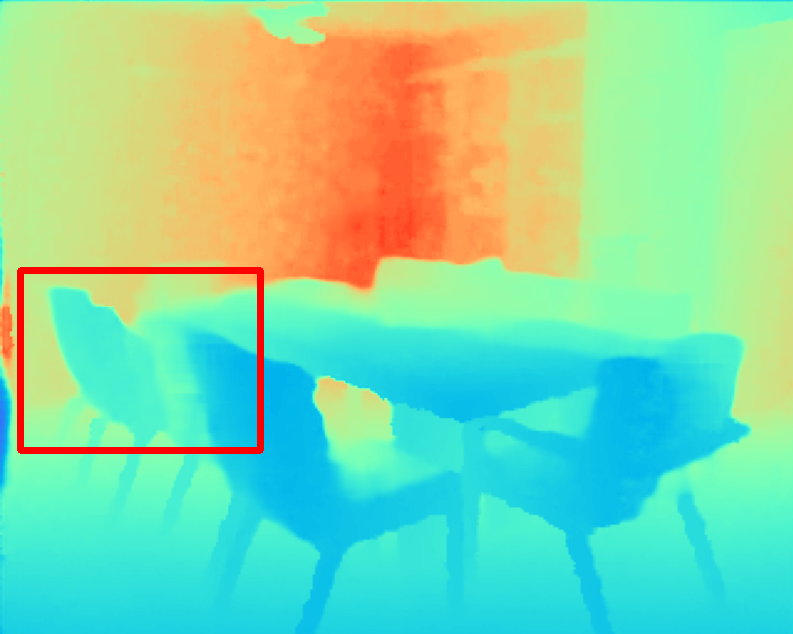}\par 
            \includegraphics[width=1\linewidth]{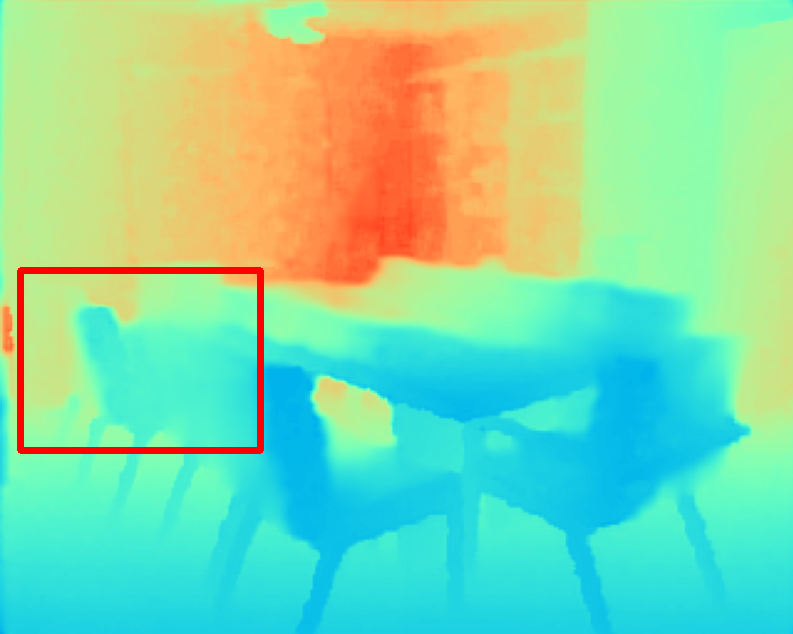}\par
            \includegraphics[width=1\linewidth]{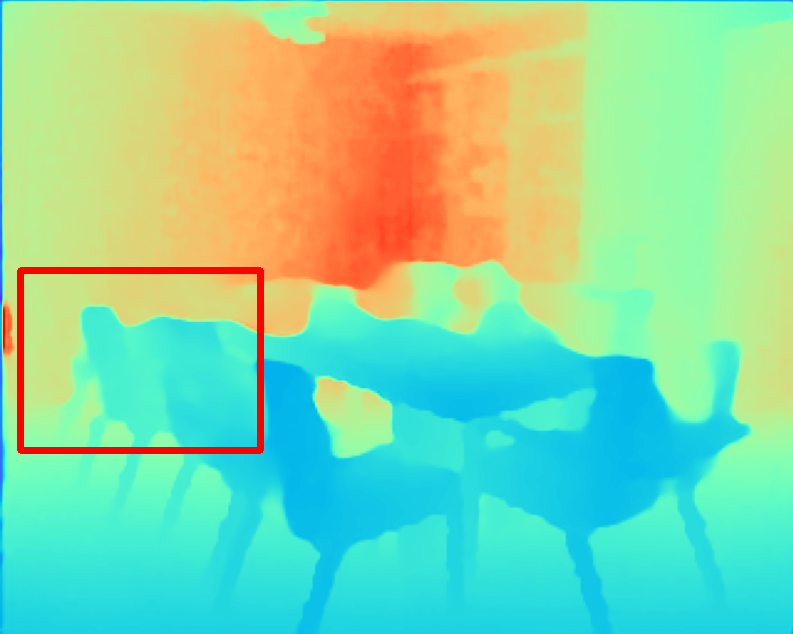}\par 
            \includegraphics[width=1\linewidth]{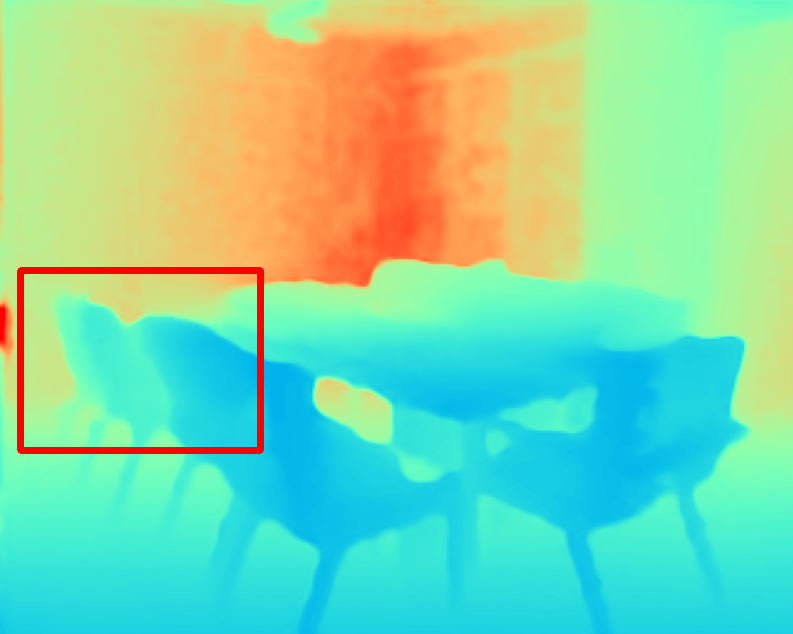}\par 
        \end{multicols}

        \begin{multicols}{7}
            \includegraphics[width=1\linewidth]{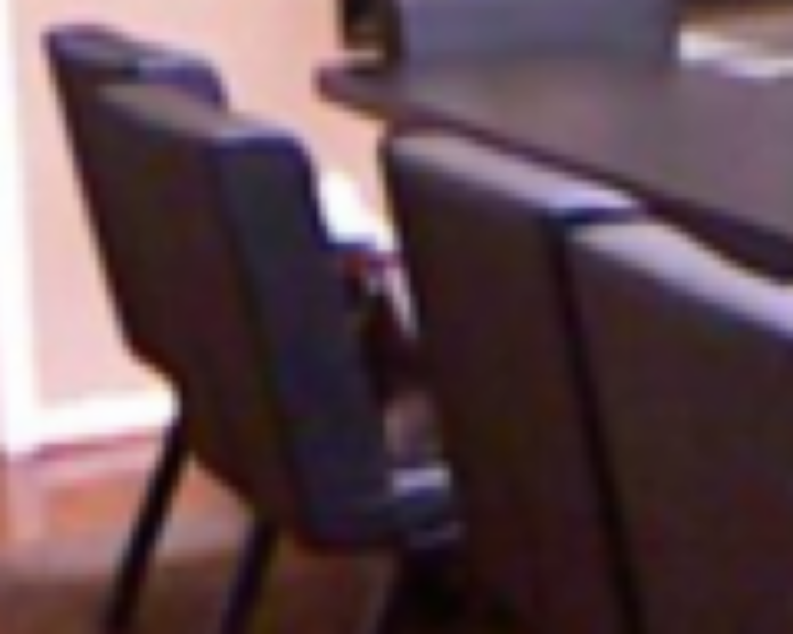}\par
            \includegraphics[width=1\linewidth]{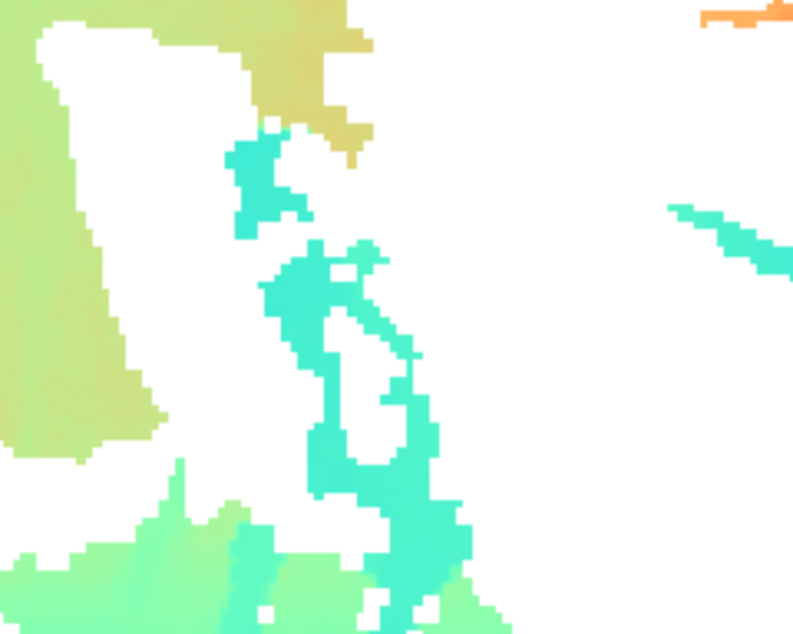}\par
            \includegraphics[width=1\linewidth]{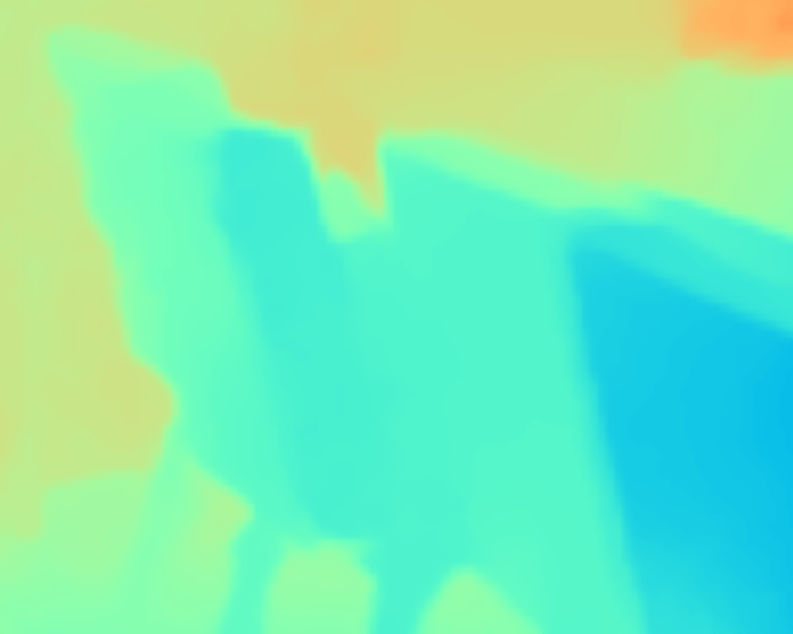}\par 
            \includegraphics[width=1\linewidth]{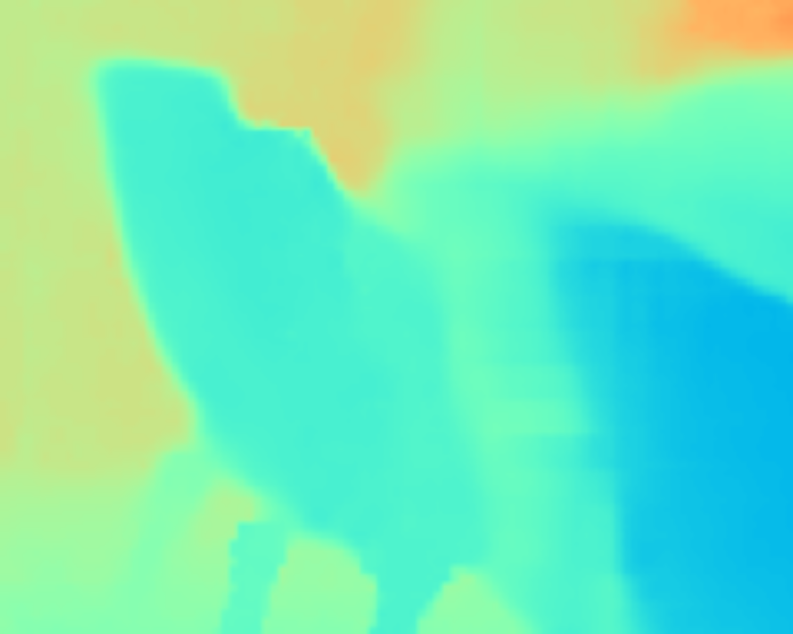}\par 
            \includegraphics[width=1\linewidth]{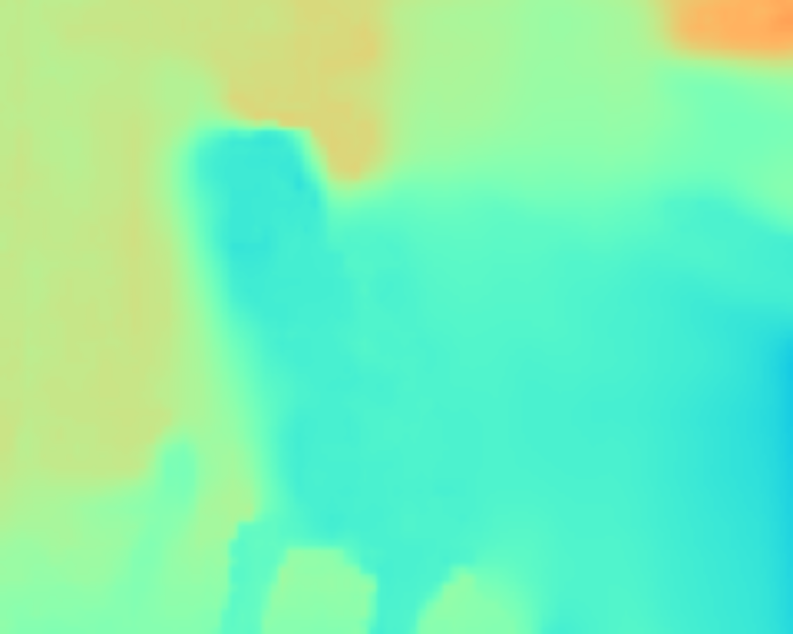}\par
            \includegraphics[width=1\linewidth]{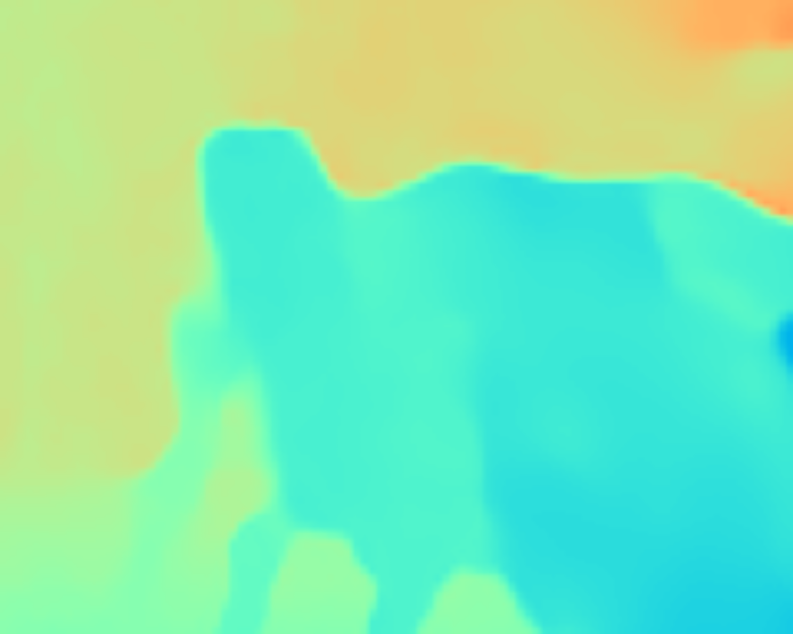}\par 
            \includegraphics[width=1\linewidth]{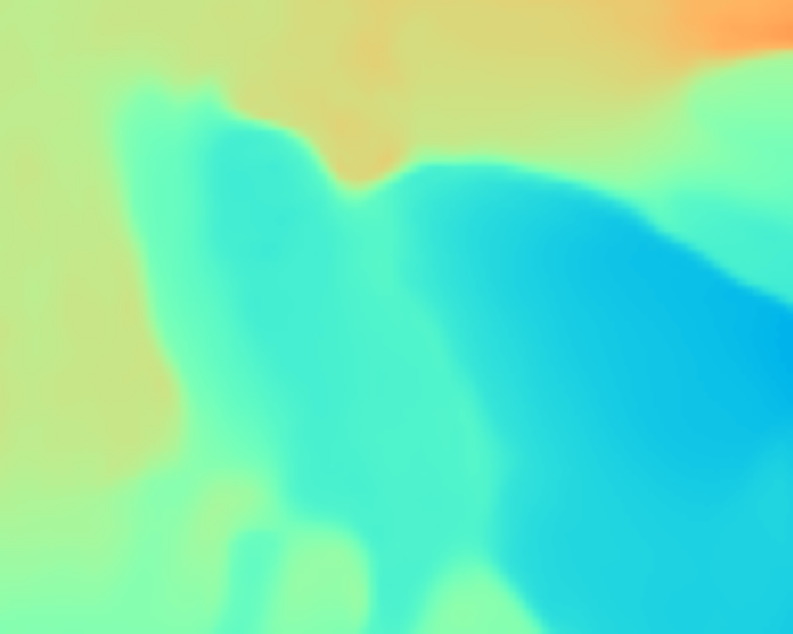}\par 
        \end{multicols}

        \begin{multicols}{7}
            \includegraphics[width=1\linewidth]{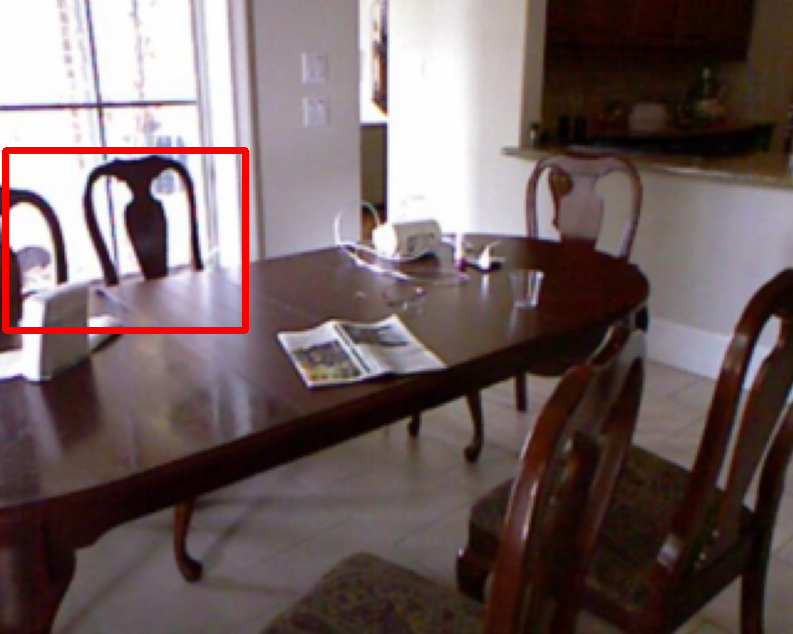}\par
            \includegraphics[width=1\linewidth]{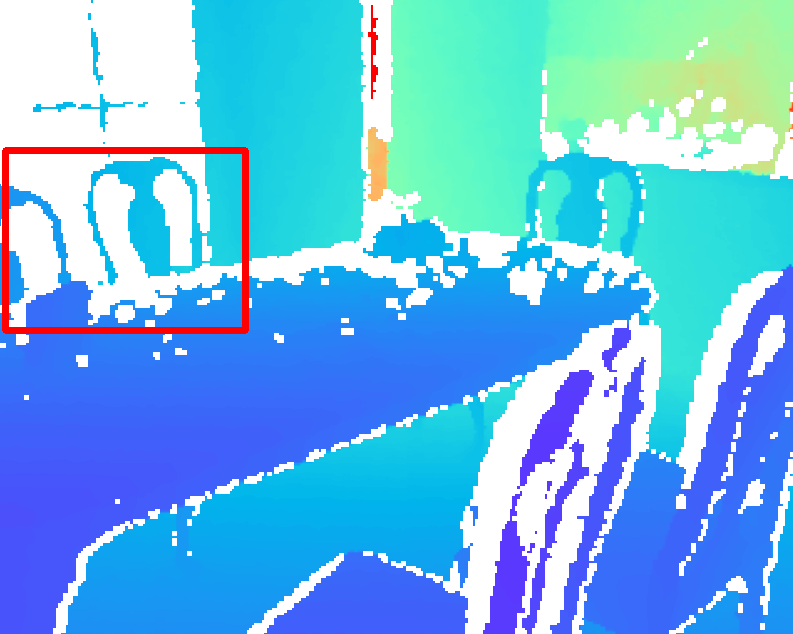}\par
            \includegraphics[width=1\linewidth]{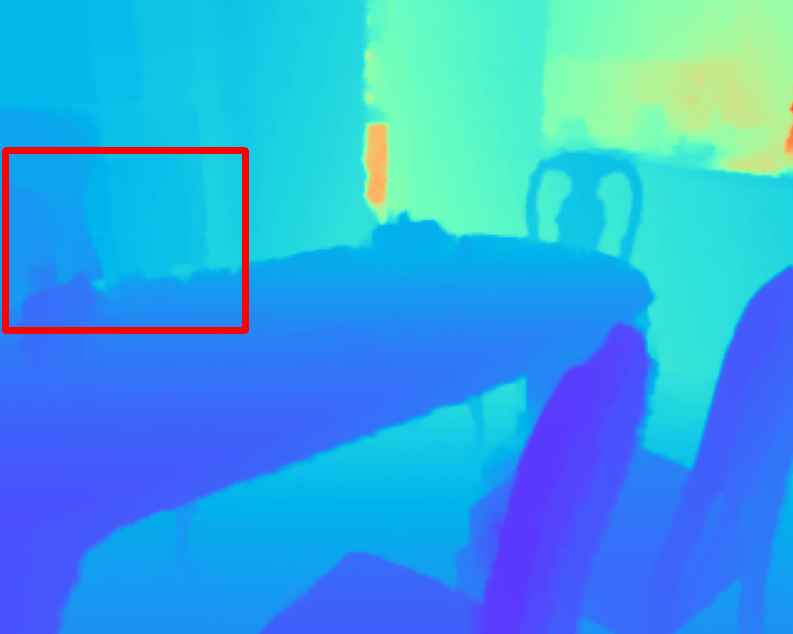}\par
            \includegraphics[width=1\linewidth]{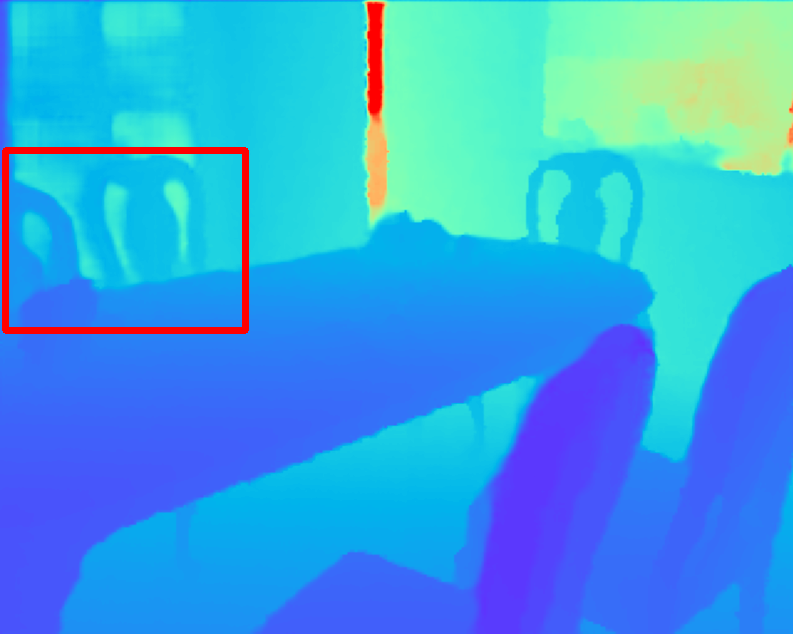}\par
            \includegraphics[width=1\linewidth]{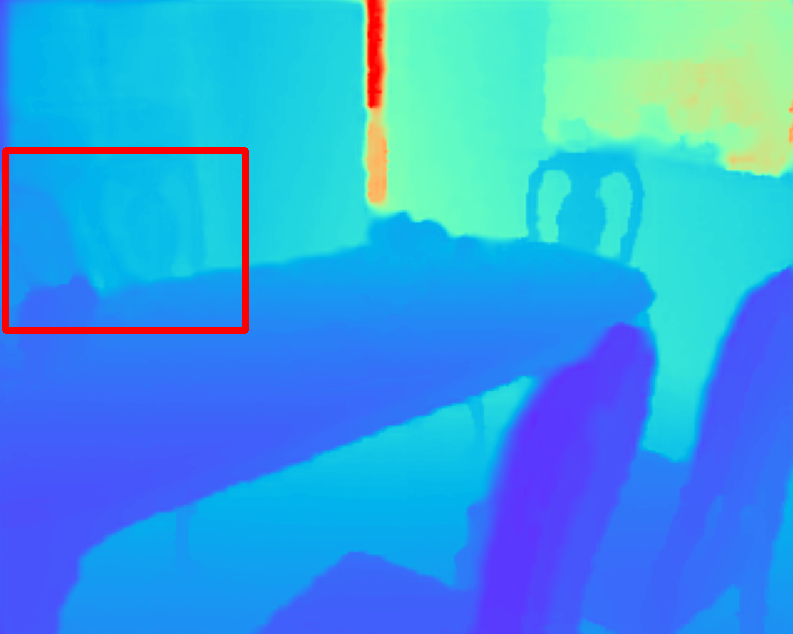}\par
            \includegraphics[width=1\linewidth]{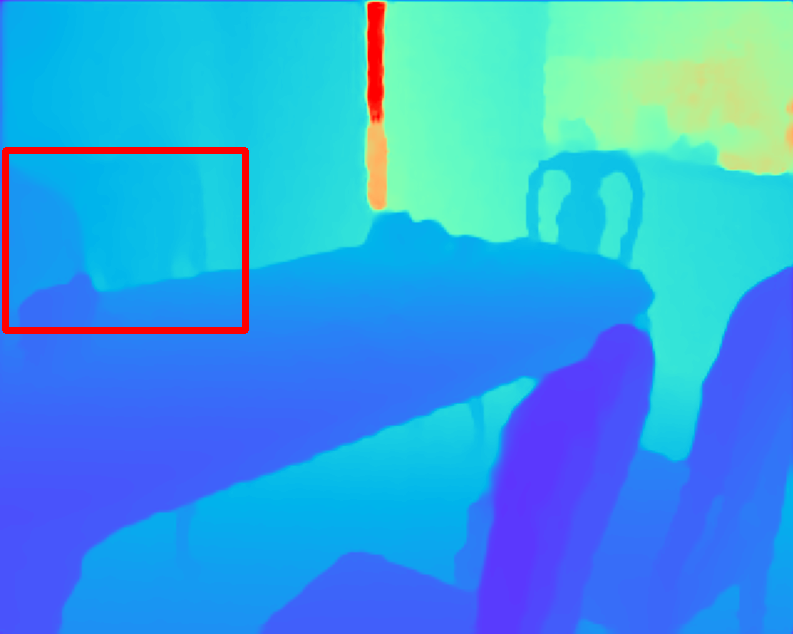}\par 
            \includegraphics[width=1\linewidth]{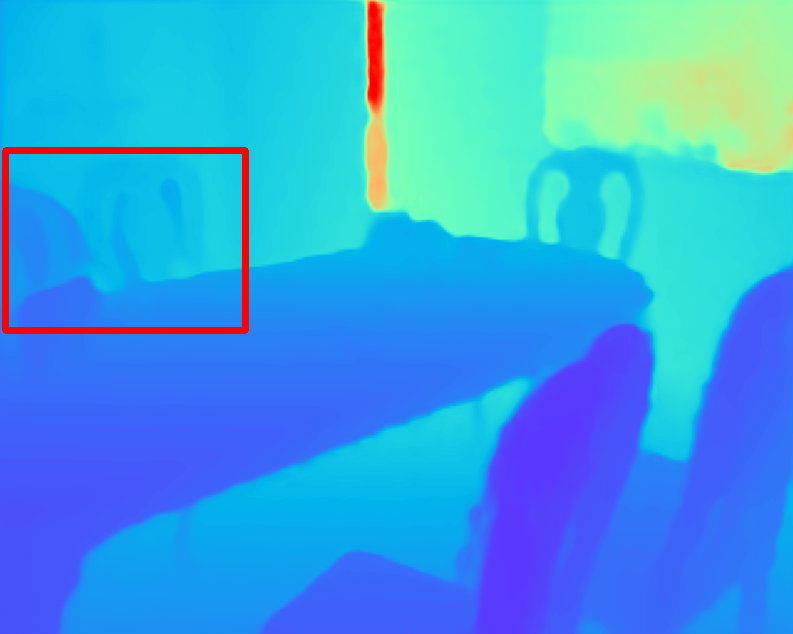}\par 
        \end{multicols}

        \begin{multicols}{7}
            \includegraphics[width=1\linewidth]{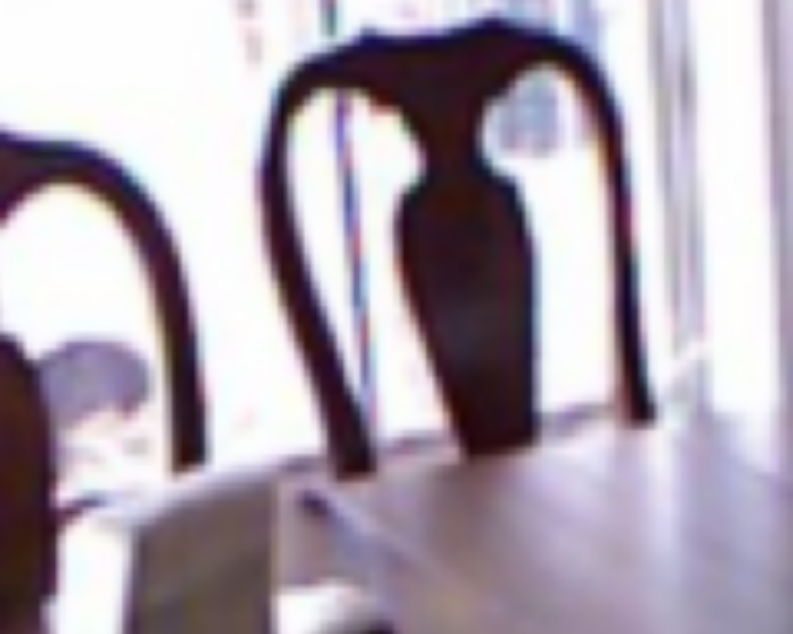} \centering \small{RGB}\par
            \includegraphics[width=1\linewidth]{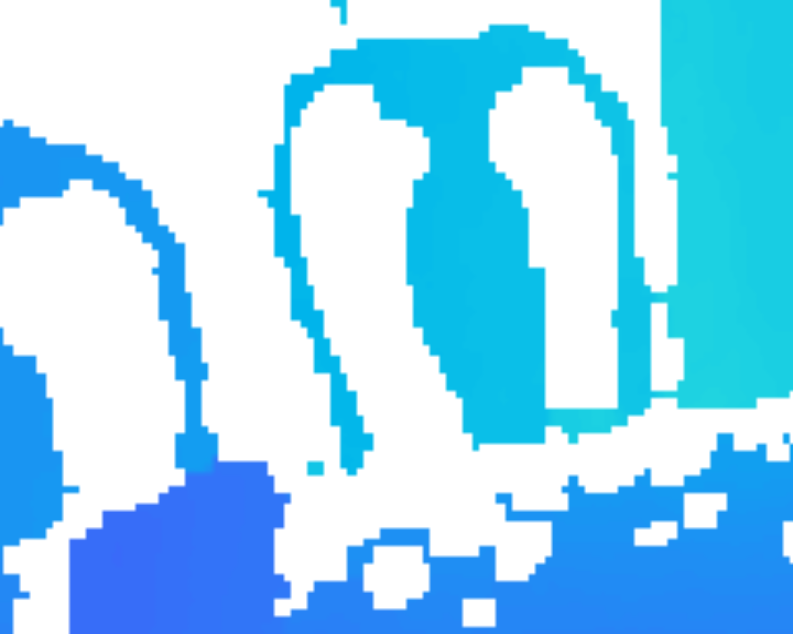} \centering \small{Sensor}\par
            \includegraphics[width=1\linewidth]{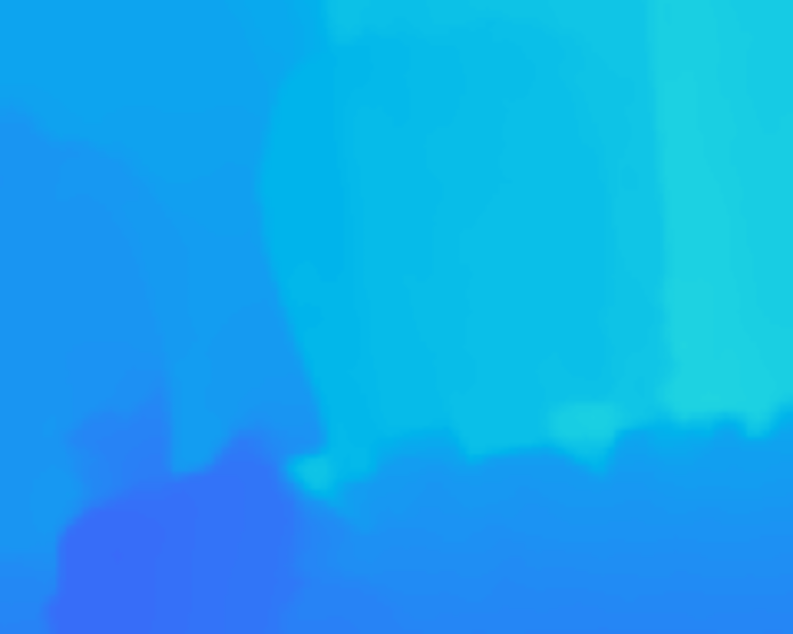} \centering \small{GT}\par
            \includegraphics[width=1\linewidth]{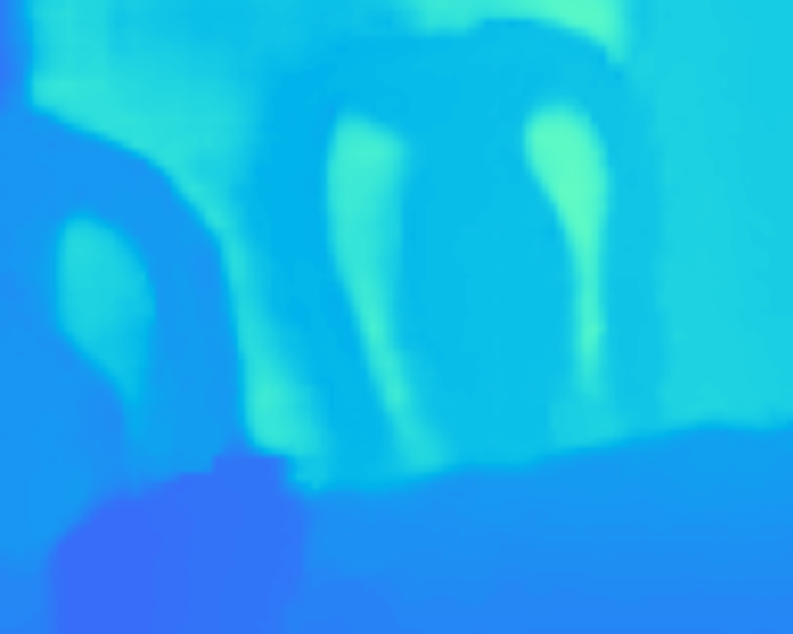} \centering \small{Gansbeke \etal \cite{wvangansbeke_depth_2019}}\par
            \includegraphics[width=1\linewidth]{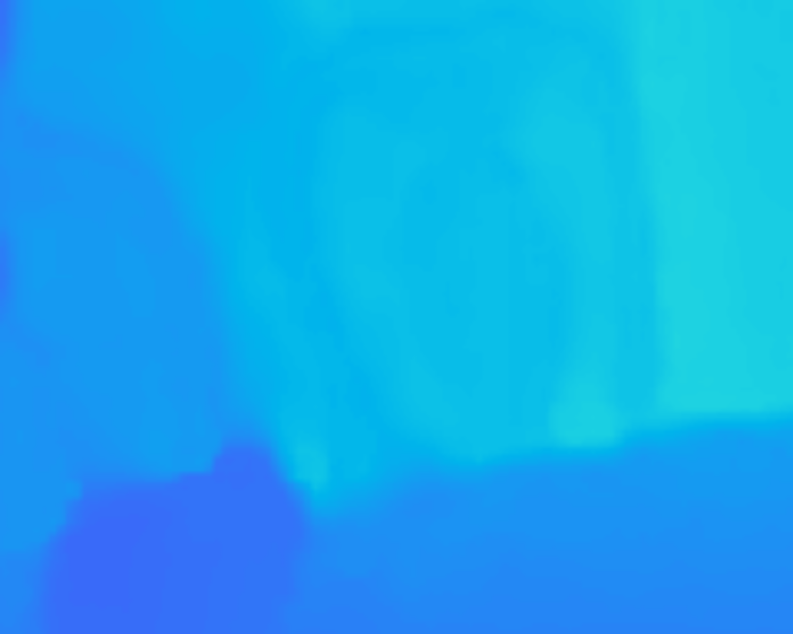} \centering \small{Li \etal \cite{msg_chn}}\par
            \includegraphics[width=1\linewidth]{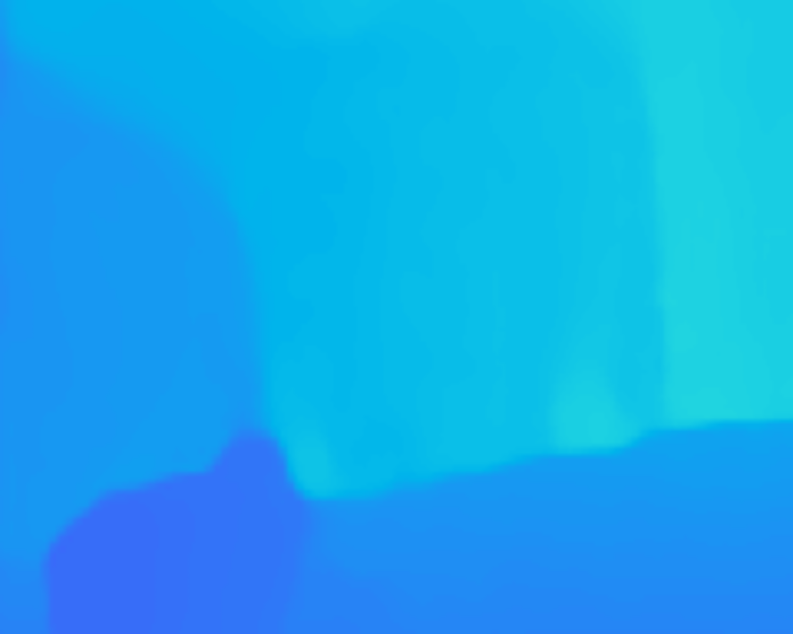} \centering \small{Huang \etal \cite{Huang_2019}}\par 
            \includegraphics[width=1\linewidth]{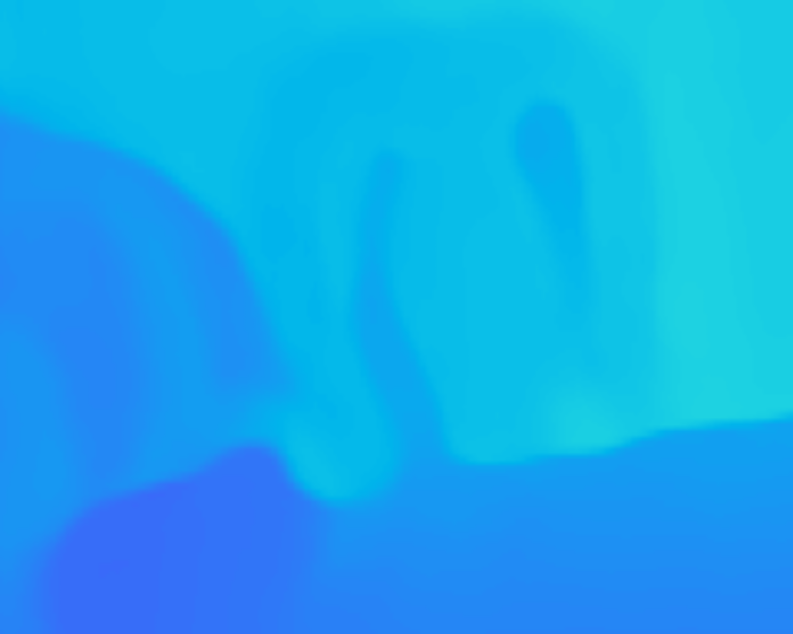} \centering \small{ours}\par 
        \end{multicols}

    \caption{Qualitative comparison with Gansbeke \etal\cite{wvangansbeke_depth_2019}, Li \etal\cite{msg_chn}, Huang \etal\cite{Huang_2019} on NYUv2 \cite{nyuv2} test set. All models are trained using our semi-dense sampling strategy. The third and fourth raws present a hard example.}
    \label{fig:nyu_viz_test}
    \end{figure*}
    
    \begin{table*}[ht]
    \setlength{\tabcolsep}{5pt}
    \renewcommand{\arraystretch}{1.0}
    \centering
	\begin{tabular}{|c|c|c|c|c|c|c|c|c|c|c|}
	    \hline
	    & \multicolumn{5}{|c}{semi-dense} & \multicolumn{5}{|c}{sparse (500 points)} \vline\\
	    \hline
	    & RMSE $\downarrow$ & rel $\downarrow$ & $\delta_{1.25}$ $\uparrow$ & $\delta_{1.25^2}$ $\uparrow$ & $\delta_{1.25^3}$ $\uparrow$ & RMSE $\downarrow$ & rel $\downarrow$ & $\delta_{1.25}$ $\uparrow$ & $\delta_{1.25^2}$ $\uparrow$ & $\delta_{1.25^3}$ $\uparrow$ \\
		\hline
		\hline
		\textbf{Huang \etal \cite{Huang_2019}} & 0.271 & 0.016 & 98.1 & 99.1 & 99.4 & -- & -- & -- & -- & -- \\
		\textbf{Gansbeke} \etal \cite{wvangansbeke_depth_2019} & 0.260 & 0.017 & 97.9 & 99.3 & 99.7 & 0.344 & 0.042 & 96.1 & 98.5 & 99.5 \\
		\textbf{Li} \etal \cite{msg_chn} & \textbf{0.190} & 0.018 & \textbf{98.8} & \textbf{99.7} & \textbf{99.9} & 0.272 & \textbf{0.034} & 97.3 & 99.2 & 99.7 \\
		\hline
		\textbf{DM-LRN} (ours) & 0.205 & \textbf{0.014} & \textbf{98.8} & 99.6 & \textbf{99.9} & \textbf{0.263} & 0.035 & \textbf{97.5} & \textbf{99.3} & \textbf{99.8} \\
		\hline
	\end{tabular}
	\vspace{0.1cm}
	\caption{\emph{NYUv2 TEST}. Quantitative comparison of training setups for different models. Semi-dense sampling preserves more accurate information that leads to better results. Although our approach is not intended to be applied to sparse depth sensors, it demonstrates strong results in the sparse training setting in indoor environments. We do not use any densification scheme for target depth reconstruction. Pseudo-sensor data is directly sampled from real sensor data.}
	\label{tab:nyu_test}
    \end{table*}
    
    \paragraph{ScanNet.} 
    In order to evaluate the generalization capability of our method, we conduct a cross-dataset evaluation. Since the test split was not provided for depth completion on \emph{ScanNet}, we use 20\% of the scenes for testing. For the sake of data diversity, we split all frames into intervals of consecutive 20 frames and take some samples out of each interval. We take the image with the largest variance of Laplacian~\cite{blurdetection} and the image with the largest file size (which indicates the level of details for a frame). We test the models trained on Matterport3D~\cite{Matterport3D} on this subset that was not seen by the models during the training process. Quantitative results are presented in Table~\ref{tab:scannet_test}, and a qualitative evaluation is shown in Fig.~\ref{fig:scannet_viz_test}. Our method provides sharp depth maps and significantly improves $\delta_{1.05}$, $\delta_{1.10}$, SSIM, and MAE metrics.
    
        \begin{figure*}[!t]
    \setlength{\tabcolsep}{1pt}
    \renewcommand{\arraystretch}{3.5}
    \vspace{-1.5cm}
    \begin{tabular}{ccc}
         &  & \\
        \rotatebox[origin=c]{90}{\small{RGB}} & 
        \includegraphics[align=c,scale=0.191]{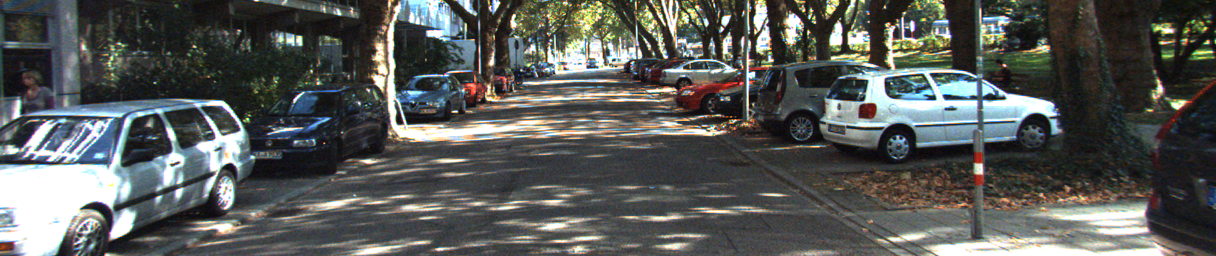} &
        \includegraphics[align=c,scale=0.191]{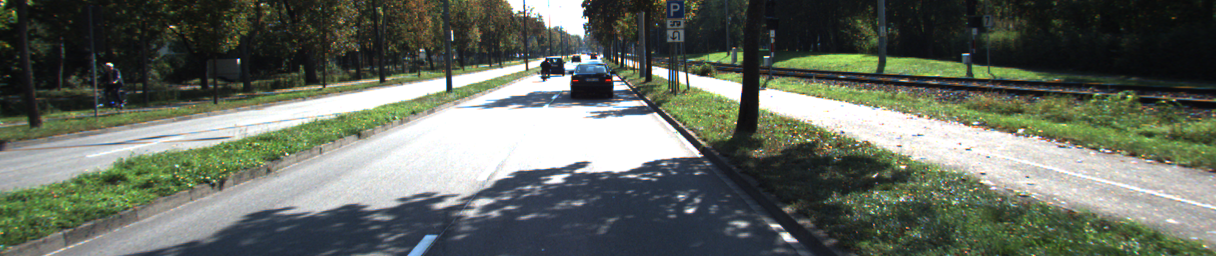}\\
        
        \rotatebox[origin=c]{90}{\small{\makecell{Tang \etal \\ \cite{guidenet}}}}&
        \includegraphics[align=c,scale=0.191]{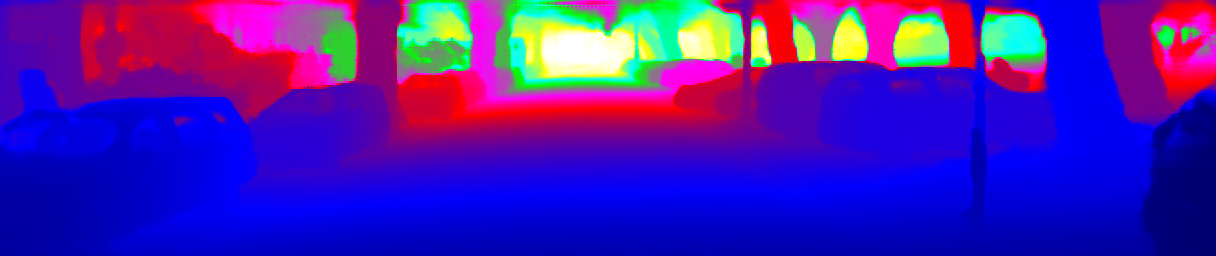}&
        \includegraphics[align=c,scale=0.191]{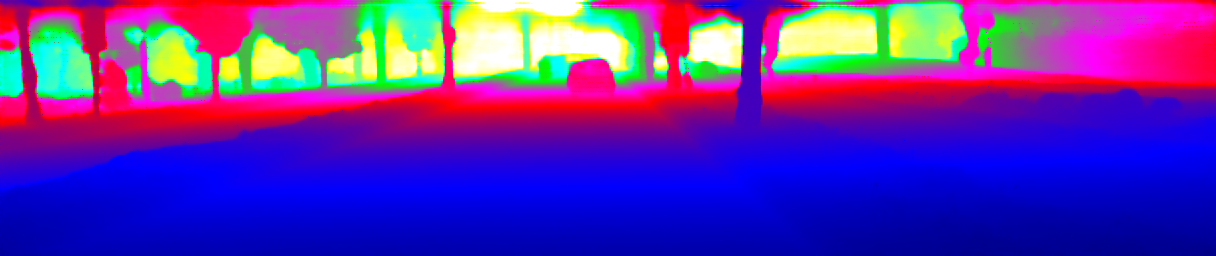}\\
        
        \rotatebox[origin=c]{90}{\small{\makecell{Li \etal \\ \cite{msg_chn}}}} &
        \includegraphics[align=c,scale=0.191]{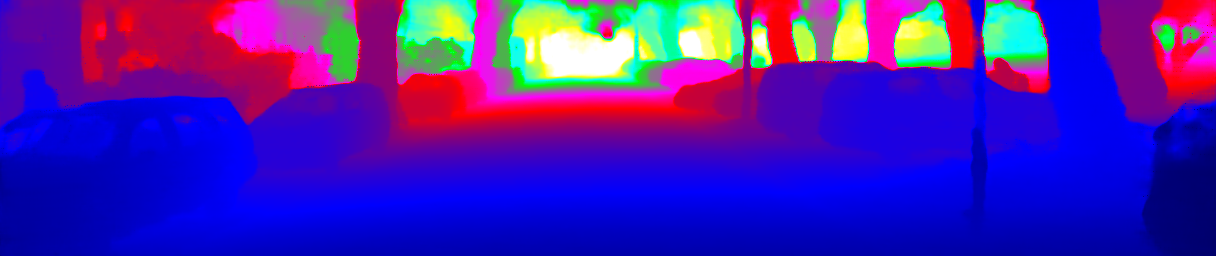} &
        \includegraphics[align=c,scale=0.191]{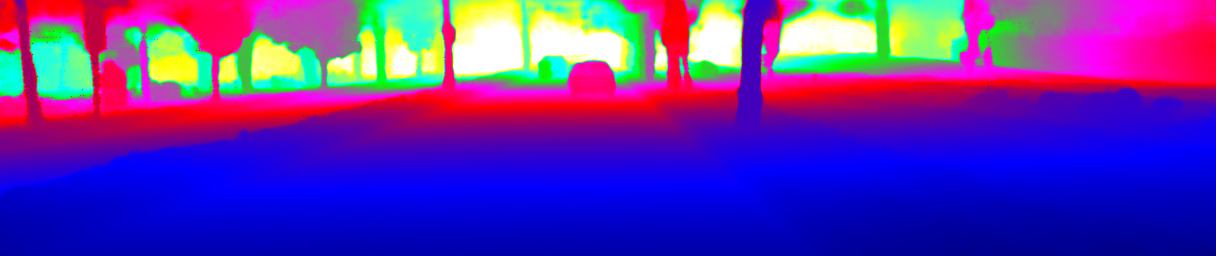} \\
        
        \rotatebox[origin=c]{90}{\small{\makecell{Gansbeke \\ \etal \cite{wvangansbeke_depth_2019}}}} &
        \includegraphics[align=c,scale=0.191]{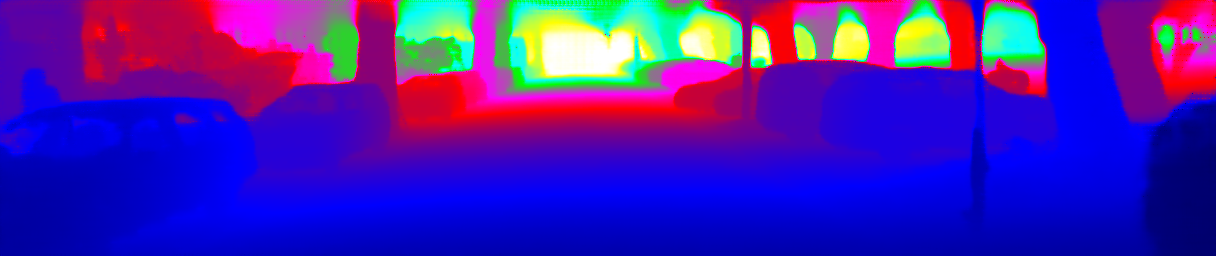} &
        \includegraphics[align=c,scale=0.191]{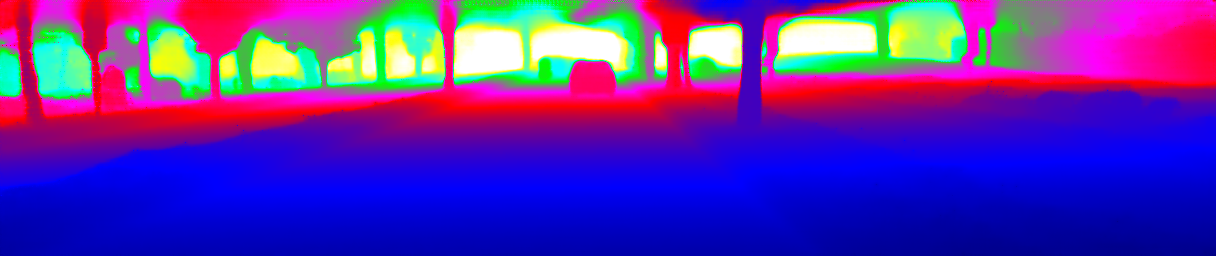} \\
        
        \rotatebox[origin=c]{90}{\small{Ours}} &
        \includegraphics[align=c,scale=0.191]{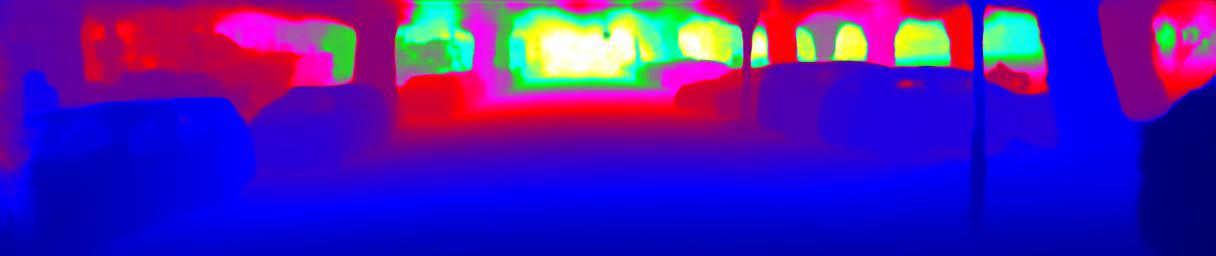} &
        \includegraphics[align=c,scale=0.191]{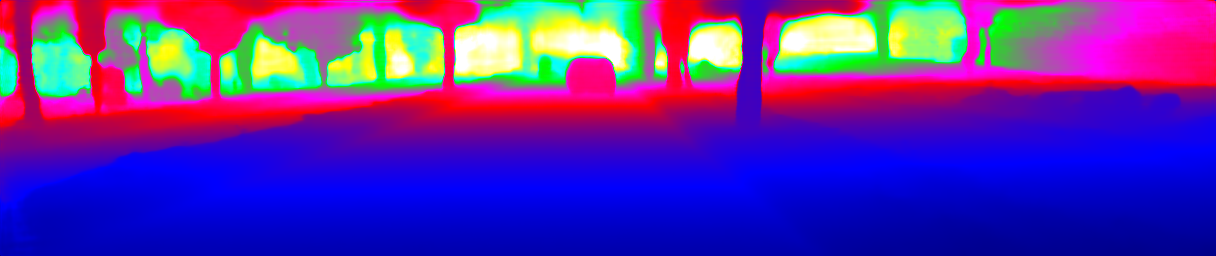} \\
    
    \end{tabular}
    \vspace{0.1cm}
    \caption{Qualitative comparison with the state-of-art methods on the KITTI test set. Even though our model was designed for a different use case scenario, it is still comparable to the best performing KITTI models in an outdoor environment.}
    \label{fig:kitti_viz_test}
    \end{figure*}
    
    \paragraph{NYUv2.} Since this dataset provides both sensor and reconstruction depth data only for the test subset, we use it to verify our training strategy that does not require ground truth. We first cut off black borders (45, 15, 45, 40 pixels from the top, bottom, left, and right side, respectively) from the original $640 \times 480$ RGBD images. Then the images are bilinearly interpolated to $320 \times 256$ resolution. These preprocessed RGBD images are used for pseudo sensor data sampling. At test time, the original sensor and ground truth depth data are used. We compare our sampling strategy with the widely used random uniform sampling approach~\cite{Ma2017SparseToDense, guidenet}. Qualitative and quantitative results are presented in Fig.~\ref{fig:nyu_viz_test} and Table~\ref{tab:nyu_test}. Since the original semi-dense depth maps contain more accurate information, our training approach demonstrates significant improvements in all target metrics. The compared performance of models originally designed for sparse inputs is shown in Table~\ref{tab:nyu_test}. Our model demonstrates strong results in this setup as well.
    
    \paragraph{KITTI.} In general, this dataset is out of our scope, since it consists of sparse LiDAR depth measurements. It is a hard case for our model, because the architecture includes a unified encoder for the joint RGBD signal, expecting \textit{segments} filled with correct depth values. Previous work~\cite{msg_chn} has demonstrated that it is a suboptimal design for a sparse depth completion model.  
    
    Since LiDAR-based outdoor depth completion differs from our use-case scenario, we perform an additional search for the most suitable loss function. As a result, we have chosen the $L_2$ loss in the logarithmic domain (see Supplementary material for more details). 
    As the LiDAR points at the top of an image are rare, input images were cropped to $256\times 1216$ for both training and testing, following~\cite{guidenet}. A horizontal flip was used as data augmentation.
    
    A quantitative comparison is shown in Table~\ref{tab:kitti_test}. Being designed for semi-dense sensors, our approach demonstrates mid-level performance compared to the KITTI leaderboard. In general, our model produces accurate depth maps, even though there are some errors at the borders of the image.
    
    \begin{table}[h]
    \setlength{\tabcolsep}{7pt}
    \renewcommand{\arraystretch}{1.0}
    \centering
    	\begin{tabular}{|c|c|c|c|c|} 
    	    \hline
    		 & RMSE & MAE & iRMSE & iMAE \\
    		\hline
    		\hline
    		Cheng \etal \cite{Cheng_2018_ECCV} & 1019 & 279 & 2.93 & 1.15  \\
    		Gansbeke \cite{wvangansbeke_depth_2019} & 773 & 215 & 2.19 & 0.93 \\
    		Lee \etal \cite{9078070} & 807 & 254 & 2.73 & 1.33 \\
    		Qiu \etal \cite{deeplidar} & 758 & 226 & 2.56 & 1.15 \\
    		Tang \etal \cite{guidenet} & 736 & 218 & 2.25 & 0.99 \\
    		Chen \etal \cite{uberfusenet}& 753 & 221 & 2.34 & 1.14 \\
    		Li \etal \cite{msg_chn} & 762 & 220 & 2.30 & 0.98 \\
    		\hline
    		Ours & 984 & 287 & 2.67 & 1.17 \\
            \hline
    	\end{tabular}
    	\vspace{0.1cm}
    	\caption{\emph{KITTI TEST}. Quantitative comparison with top ranked KITTI models. All metrics are measured in millimeters.}
    	\label{tab:kitti_test}
    \end{table}

\section{Conclusion}\label{sec:concl}

In this work, we have proposed a new depth completion method for semi-dense depth sensor maps with auxiliary color images. Our main innovation is a novel decoder architecture that exploits statistical differences between mostly filled and mostly empty regions. It is implemented by an additional decoder modulation branch that takes a mask of missing values as input and adjusts the activation mask distribution in the decoder via SPADE blocks.

In experimental evaluation, our model has shown state-of-the-art results on the \emph{Matterport3D} dataset with generalization to ScanNet, and even competitive performance on the KITTI dataset with sparse depth measurements. We have also proposed a new training strategy for datasets with raw sensor data and without reconstructed ground truth depth, which allows us to achieve strong results on the NYUv2 dataset.

{\small
\bibliographystyle{ieee_fullname}
\bibliography{cvpr}
}

\section*{Appendix A: Loss function ablation study.}

    Firstly, we investigate how the choice of the loss function affects the performance in different use cases. We search for an appropriate loss function among the popular single-term loss functions that include $L_1$ and $L_2$ penalties in depth and log-depth domains and their pairwise variations. $L_1$ loss family appears to be more efficient for indoor semi-dense depth completion. These functions provide a balance between RMSE and MAE and improve accurate $\delta$-metrics. In other words, they produce clearer edges and boundaries. The pairwise log-$L_1$ appears to be the most suitable for Matterport3D. More details can be found in Table \ref{tab:mp3d_loss_study}.
    
    The same search procedure performed on KITTI validation set reveals an advantage of $L_2$-family. A good balance was achieved by using these penalties. Even though we choose the log-$L_2$ as the main penalty for outdoor LiDAR-oriented depth completion, it can be switched by its pairwise counterpart. A quantitative comparison is shown in Table \ref{tab:kitti_loss_study}.

    \begin{table}[ht]
    \setlength{\tabcolsep}{5pt}
    \renewcommand{\arraystretch}{1.1}
    \centering
    	\begin{tabular}{ c|c|c|c|c|c} 
    		 & RMSE & MAE & $\delta_{1.25}$ & $\delta_{1.25^2}$ & $\delta_{1.25^3}$\\
    		\hline
    		\hline
    		$l_1$ & 1.001 & 0.288 & 0.888 & 0.931 & 0.949  \\
    		$l_2$ & 0.995 & 0.311 & 0.859 & 0.924 & 0.948 \\
    		$\log l_1$ & 1.001 & 0.289 & 0.889 & 0.930 & 0.948 \\
    		$\log l_2$ & 1.006 & 0.318 & 0.869 & 0.928 & 0.949 \\
    		pairwise $\log l_1$& \textbf{0.961} & 0.285 & 0.890 & 0.933 & 0.949 \\
    		pairwise $\log l_2$& 1.020 & 0.337 & 0.859 & 0.922 & 0.947 \\
    		\hline
    
    	\end{tabular}
    	\vspace{0.1cm}
    	\caption{\emph{Matterport3D TEST}. Quantitative comparison of the popular single-term loss functions for depth estimation / completion. RMSE and MAE are measured in meters.}
    	\label{tab:mp3d_loss_study}
    \end{table}

    \begin{table}[ht]
    \setlength{\tabcolsep}{7pt}
    \renewcommand{\arraystretch}{1.1}
    \centering
    	\begin{tabular}{ c|c|c|c|c}    
    		 & RMSE & MAE & iRMSE & iMAE \\
    		\hline
    		\hline
    		$l_1$ & 1107 & 279 & 2.98 & 1.11  \\
    		$l_2$ & 1053 & 304 & 3.18 & 1.40 \\
    		$\log l_1$ & 1108 & 295 & 2.89 & 1.15 \\
    		$\log l_2$ & \textbf{1040} & 289 & 2.73 & 1.15 \\
    		pairwise $\log l_1$& 1104 & 280 & 2.88 & 1.08 \\
    		pairwise $\log l_2$& 1054 & 279 & 2.69 & 1.07 \\
    		\hline
    
    	\end{tabular}
    	\vspace{0.1cm}
    	\caption{\emph{KITTI VALIDATION}. Quantitative comparison of the popular single-term loss functions for depth estimation / completion. All metrics are measured in millimeters.}
    	\label{tab:kitti_loss_study}
    \end{table}

\section*{Appendix B: Backbone Depth.}
    
\begin{figure}[!h]
        \centering
        \includegraphics[width=1.0\linewidth]{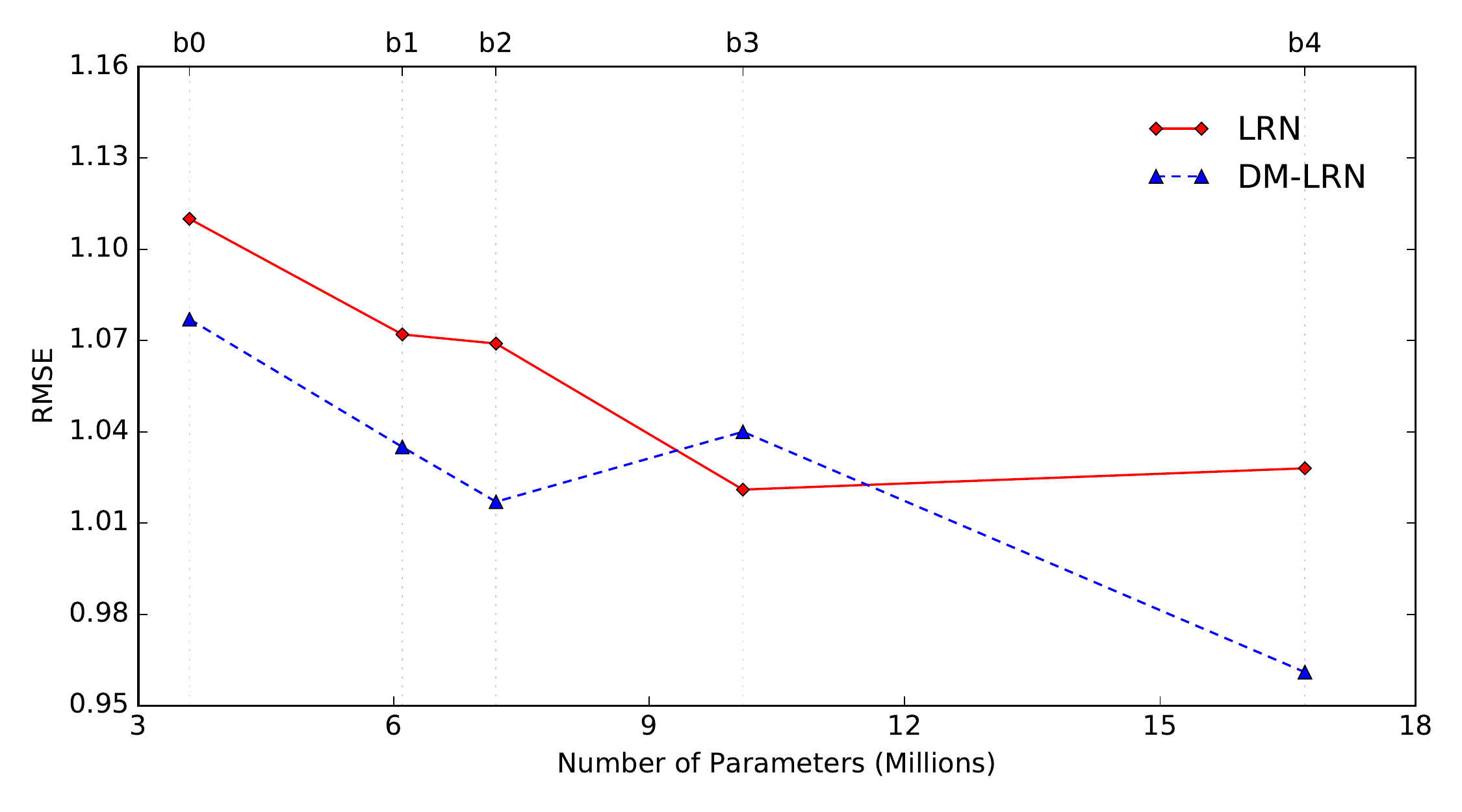}
        \caption{\emph{Matterport3D TEST}. A dependency of RMSE of the baseline model and the model with the decoder modulation concerning the size of the backbone. LRN is the baseline model with RGBD inputs. DM-LRN is the baseline with the decoder modulation branch.}
        \label{fig:backbone_study}
    \end{figure}

    Backbone scalability is an advantage of our approach. In order to investigate the behavior of the model, we carried out additional experiments in which we tried EfficientNets of different sizes. The results are shown in Figure \ref{fig:backbone_study}. The general trend is predictable: as the size of the network grows, the validation error drops. This is true for the model with and without modulation. The mask modulation consistently gives an improvement in the target metric with the exception "B3" configuration that demonstrated an unexpected behavior, assumed to be a random outlier. In order to comply with practical applications, we did not try the configurations larger than "B4".
    
\end{document}